\pdfoutput=1
\documentclass[11pt]{article}

\usepackage[final]{acl}

\usepackage{times}
\usepackage{latexsym}

\usepackage[T1]{fontenc}
\usepackage[utf8]{inputenc}

\usepackage{microtype}

\usepackage{inconsolata}

\usepackage{graphicx}

\usepackage{float}
\usepackage{multirow}
\usepackage{phoenician}
\usepackage{devanagari}
\usepackage{multicol}
\usepackage{subcaption}

\usepackage{xspace}
\usepackage{xcolor}
\newif\ifdraft\draftfalse% set to \draftfalse to hide comments
\newif\ifsub\subfalse
\definecolor{wellesley}{RGB}{0, 39, 118}

\newcommand{\aw}[1]{\ifdraft{\color{wellesley}[ #1 -- Arya]}\fi}

\newcommand{\style}{visual style\xspace}
\newcommand{\styles}{visual styles\xspace}
\newcommand{\benchmark}{GlyphPattern\xspace}
\newcommand{\gpt}{GPT-4o\xspace}
\newcommand{\gemini}{Gemini-1.5\xspace}
\newcommand{\idefics}{Idefics2\xspace}
\newcommand{\ideficsN}{Idefics3\xspace}
\newcommand{\blip}{InstructBLIP\xspace}
\newcommand{\llava}{LLaVA-NeXT\xspace}
\newcommand{\molmo}{Molmo\xspace}
\newcommand{\molmoo}{Molmo-O\xspace}
\newcommand{\molmod}{Molmo-D\xspace}
\usepackage{enumitem}
\setitemize{itemsep=0pt}

\title{\benchmark: An Abstract Pattern Recognition Benchmark for Vision-Language Models}

  \author{Zixuan Wu \\ Northeastern University \\  zi.wu@northeastern.edu \\
        \And  Yoolim Kim \\ Wellesley College \\ ykim6@wellesley.edu \\
        \And Carolyn Jane Anderson \\ Wellesley College \\
         carolyn.anderson@wellesley.edu}

\begin{document}
\maketitle
\begin{abstract}
Vision-Language Models (VLMs) have made rapid progress in reasoning across visual and textual data. While VLMs perform well on vision tasks that they are trained on, our results highlight key challenges in abstract pattern recognition. We present \benchmark, a 954 item dataset that pairs 318 human-written descriptions of visual patterns from 40 writing systems with three visual presentation styles. 

\benchmark evaluates abstract pattern recognition in VLMs, requiring models to understand and judge natural language descriptions of visual patterns. \benchmark patterns are drawn from a large-scale cognitive science investigation of human writing systems; as a result, they are rich in spatial reference and compositionality. Our experiments show that \benchmark is challenging for state-of-the-art VLMs (\gpt achieves only 55\% accuracy), with marginal gains from few-shot prompting. Our detailed analysis reveals errors at multiple levels, including visual processing, natural language understanding, and pattern generalization.
\end{abstract}

\section{Introduction}

A key aspect of intelligence is the ability to recognize abstract patterns. \citet{bongard} proposed a task to measure abstract pattern recognition in artificial intelligence using a kind of partitioned diagram, now known as a \textit{Bongard problem}. Each diagram shows two sets of items; the goal is to describe the pattern that distinguishes them. Bongard problems have inspired a number of visual benchmarks~\citep{Nie2020Bongard,jiang_bongardHOI_2022,spratley_unicode_2023,wu2024bongardopen}. However, existing benchmarks are easier than classical Bongard problems because they do not involve a natural language description of the pattern.

We present \benchmark, an abstract pattern recognition benchmark for vision-language models (VLMs) that challenges models to understand natural language pattern descriptions. \benchmark tests the ability of VLMs to recognize visual properties that distinguish two groups of symbols, pairing together human-written and human-verified patterns with three different \styles designed to test the robustness of abstract pattern recognition in VLMs, for a benchmark of 954 items. %Our results show that this is a challenging benchmark, even for state-of-the-art VLMs: GPT-4o achieves only X accuracy on the multiple choice task, and Y correctness in free response pattern generation.

\begin{figure}
    \includegraphics[width=0.85\linewidth]{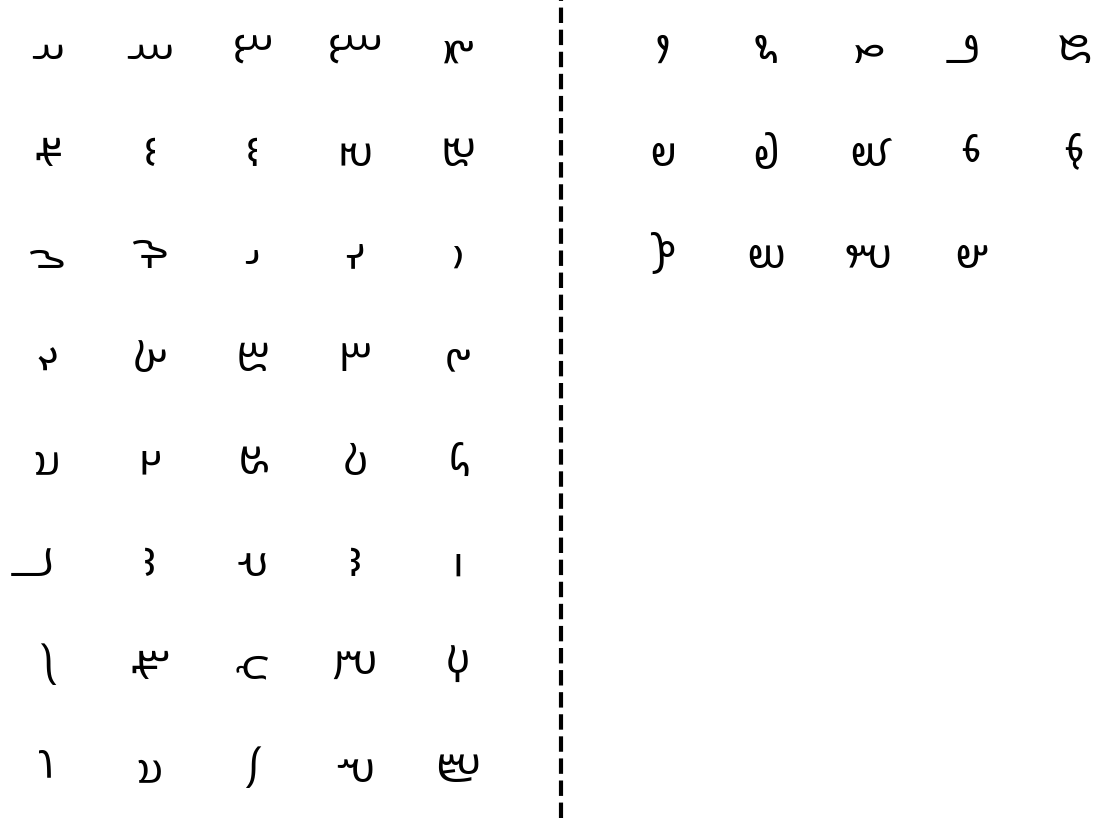}
    
    \textbf{Characters on the right side in the image are characters that contain a loop or circular shape.}
    \caption{Example \benchmark item in the left-right \style. The writing system is Avestan.}
    \label{fig:ex}
\end{figure}

A key feature of \benchmark is that it builds upon the Glyph project~\citep{glyph}, a cognitive science investigation of cross-linguistic universals in human writing systems that collects human-written patterns in a gamified web-based experiment. Glyph provides a high quality data source since each rule is validated by multiple players. 

\begin{figure*}[t]
    \begin{multicols}{2}
    \begin{subfigure}{\linewidth}
    \centering
    \fbox{\includegraphics[width=0.8\linewidth]{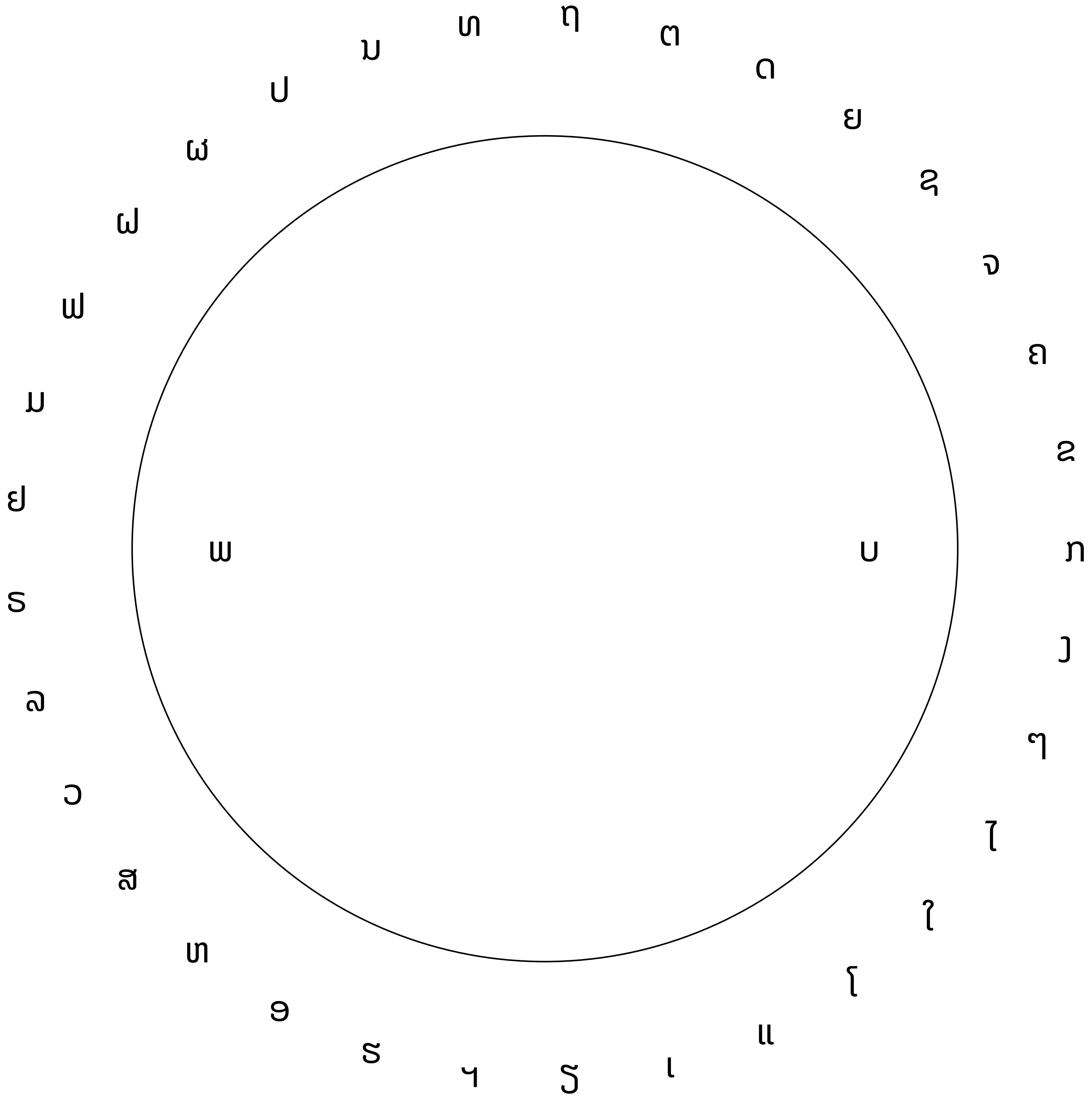}}
    \caption{Circular partition style.}
    \end{subfigure}
    
    \begin{subfigure}{\linewidth}
    \centering
    \fbox{\includegraphics[width=0.8\linewidth]{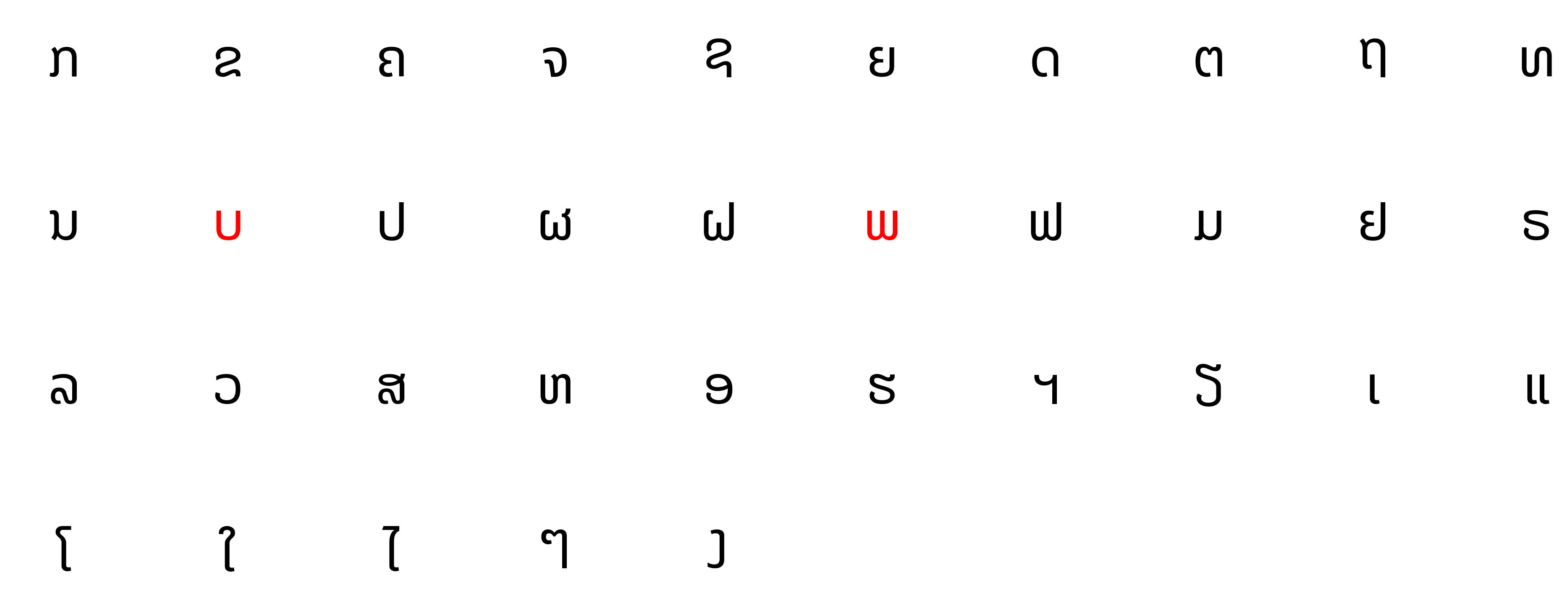}}
    \caption{Color contrast style.}
    \fbox{\includegraphics[width=0.55\linewidth]{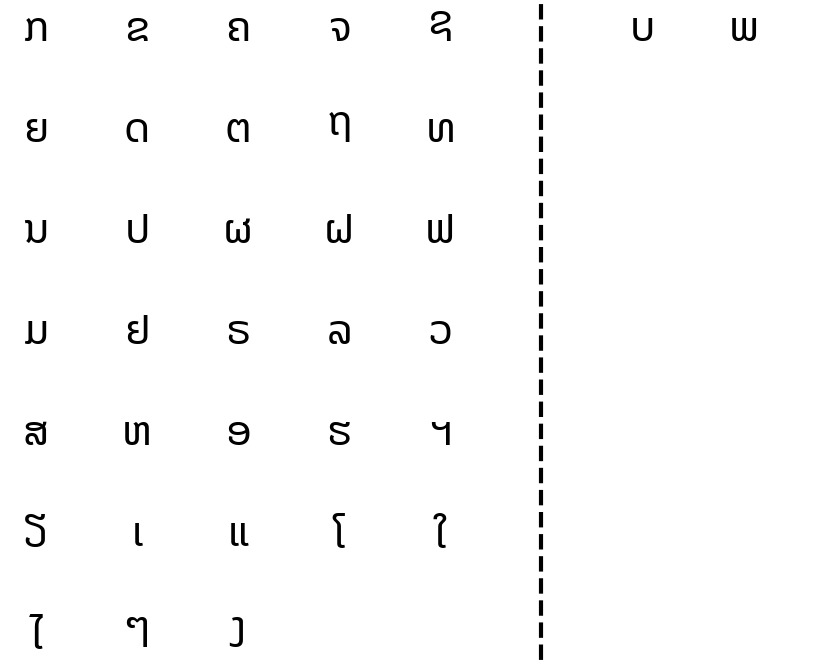}}
    \caption{Left-right partition style.}
    \end{subfigure}
    \end{multicols}
    \caption{\benchmark's three \styles for the Lao pattern \textit{Characters that are symmetrical around a vertical axis.}}
    \label{fig:image_ex}
\end{figure*}

Figure \ref{fig:ex} shows an example item from \benchmark. Each human-written rule describes a visual property of a subset of the characters in the writing system: in this case, the characters on the right of the image contain a loop, while the characters on the left do not. We test three different \styles: a left-right partition, a circular boundary, and a color contrast (Figure \ref{fig:image_ex}). 

We use \benchmark to test the pattern recognition ability of eight VLMs in a multiple choice task. Our zero-shot results show that this is a challenging task for all models: the state-of-the-art \gpt model achieves only 55.6\% accuracy. We find that although few-shot prompting does not help \gpt, it enables \gemini to match \gpt's performance. For the best-performing models, we also explore a free response format and find that models generate valid rules less than 16\% of the time.

%The Glyph project uses a gamified web-based experiment to collect human descriptions of rules that distinguish a subset of characters in a writing system. 
 %We extract 318 human-identified visual patterns from the publicly available Glyph data, spanning 40 writing systems. 
\benchmark's rich human-written pattern set and multiple \styles allow us to explore the impact of different aspects of the task, including the length and syntactic complexity of the rules, the visual saliency of different writing systems, and the visual presentation style. Our fine-grained error analysis reveals that models ``fail to see'' key portions of the image and fail to understand certain terms in the natural language descriptions. We hope that \benchmark will guide improvements in VLMs by highlighting weaknesses in both the vision and language capabilities of current models. Our key contributions are as follows:

\begin{itemize}
    \item A 954 item abstract pattern recognition benchmark for VLMs based on human-written descriptions of patterns in visual properties of writing system characters.
    \item Benchmarking results for eight VLMs, showing that \benchmark is challenging for state-of-the-art models: \gemini and \gpt achieve accuracy around 55\%.
    \item A comparison of zero-shot, few-shot, chain-of-thought, and free response results for the two best-performing models.
    \item A detailed analysis of the impact of image style, pattern complexity, and human-identified cross-linguistic pattern types.
    \item An exploration of error sources, revealing mistakes that arise at multiple stages, including image-processing, language understanding, and abstract pattern reasoning.
\end{itemize}

\section{Related Work} 

There is a long history of assessing intelligence through the ability to recognize abstract patterns, both in non-verbal tests of human intelligence and in AI. \citet{bongard}'s influential work on pattern recognition in AI proposed a test using a diagram that partitions items into two groups. So-called \textit{Bongard problems} evaluate an agent's few-shot pattern recognition ability: the goal is to recognize and describe the pattern from a handful of examples. 

\subsection{Abstract Pattern Recognition Benchmarks}

General-purpose visual understanding requires the ability to recognize patterns from a handful of examples. Consequently, it is important to understand the abstract reasoning abilities of VLMs alongside their real-world, domain-specific performance~\citep{hudson_gqa_2019,mmmu}.

A number of abstract pattern recognition benchmarks for few-shot image classification have been inspired by intelligence tests. \citet{barrett2018measuringabstractreasoningneural} adapt IQ test items to measure few-shot pattern recognition in computer vision models. The model is prompted with a grid of example images, and asked to select the next member of the sequence. \citet{zhang2019raven} and \citet{teney2020v} use a similar format inspired by Raven’s Progressive Matrices~\citep{raven}. \citet{zerroug2022benchmark}, by contrast, use an odd-one-out classification task. 

There are several Bongard-inspired image classification benchmarks. \citet{Nie2020Bongard} propose Bongard-Logo, a benchmark where models classify images as members or non-members of a pattern. Bongard-HOI~\cite{jiang_bongardHOI_2022} and Bongard-OpenWorld~\cite{wu2024bongardopen} use real-world images. Most similar to our work, \citet{spratley_unicode_2023} present a dataset of Unicode character patterns; unlike our patterns, which are easily solved by humans, their automatically-identified patterns have only a 55\% human solve rate. 

These benchmarks are a simplification of Bongard's original task, because they test whether the model can \textit{continue} a pattern rather than \textit{describe} it. Our \benchmark benchmark comes closer by challenging VLMs to understand and even produce natural language descriptions of abstract patterns.

\subsection{Benchmarking Vision-Language Models}

Our focus is on benchmarking VLMs, which may struggle with image processing, natural language understanding, or relating the two modalities. 

\paragraph{Compositionality in VLMs} An important feature of \benchmark is that its human-written patterns are linguistically rich in compositionality and spatial relations. Compositionality is a key evaluation area for VLMs. Existing benchmarks explore compositionality in object attributes~\cite{ray2023cola}, spatial relations~\citep{liu-etal-2023-visual,kamath-etal-2023-whats}, and visual concepts~\cite{suhr-etal-2019-corpus,ma2023crepe}. A popular benchmark is Winoground~\cite{thrush2022winoground}, which targets linguistic compositionality with pairs of captions containing the same words in different syntactic structures. \citet{diwan-etal-2022-winoground} argue that it is text-image linking that makes Winoground challenging. 

\paragraph{Abstract Pattern Recognition in VLMs}

The increasing prominence of LLM-based systems has led to widespread interest in multimodality. However, recent work on abstract pattern recognition reveals surprising weaknesses of VLMs. \citet{ji-etal-2022-abstract} find poor zero-shot performance on tangram puzzle-based tasks. \aw{\citet{chia2024puzzlevqa} task VLMs to solve abstract puzzles, and find that model struggle with visual perception, inductive reasoning, and deductive reasoning abilities.}\citet{rahmanzadehgervi2024vision} show that VLMs struggle with extremely easy visual reasoning tasks, like judging if circles overlap, which they attribute to visual processing issues. 
\citet{wu2024surprising} ask VLMs to verify captions for displays of geometric shapes from the NLVR dataset \cite{suhr-etal-2017-corpus}, and find that even state-of-the-art models perform little better than randomly. 

Our findings converge with and expand on \citet{wu2024surprising}'s findings: using a different task and data domain, we find similar weaknesses, including inconsistent gains from few-shot prompting. However, our fine-grained analysis allows us to dig deeper into the sources of errors. 

\paragraph{Computer Vision and Writing Systems}

\benchmark items come from a diverse set of human writing systems. There is previous work on computer vision systems for recognizing characters in writing systems, including \citet{magre2022typography}'s MNIST~\citep{deng2012mnist} inspired dataset and \citet{lake2015human}'s work on discovering schemas for recognizing characters. Our goal is to probe abstract pattern recognition in VLMs, rather than to improve their ability to transcribe writing systems.

\section{Dataset}\label{sec:dataset}

\benchmark uses patterns from human writing systems to benchmark abstract pattern recognition in VLMs. It pairs 318 human-written descriptions of patterns that partition characters in 40 human writing systems with three different visual presentation styles, for a total of 954 items.\footnote{The dataset and support code are available at \url{https://github.com/Wellesley-EASEL-lab/GlyphPattern}.}

\subsection{Online Data Collection}

We draw on human-written patterns collected as part of the online applet \textit{Glyph} created by the Minds and Traditions Max Planck Independent Research Group, which explores cross-linguistic universals in the visual properties of human writing systems. We use the publicly-available data collected from the online applet. 

In \textit{Glyph}, players select a subset of characters in an unfamiliar writing system and write a description of the pattern that separates the selected characters (the \textit{included characters}) from the rest of the script (the \textit{excluded characters}). Players must verify their rule by re-selecting the included characters after at least three minutes using their previously written rule.

\subsection{Data Cleaning}

We use a subset of 318 filtered, validated, and standardized player-written rules produced by the \textit{Glyph} team. Rules were independently validated by five researchers, who identified redundant rules and grouped the rules into categories. Rules were then summarized by ChatGPT in order to standardize wording. We use this publicly-available subset of rules, but have an additional rewording step so that all rules are compatible with our prompt prefixes.

\subsection{Image Construction}

The Glyph dataset associates each rule with a script and a partition of characters. We retrieve the Unicode symbols for the included and excluded characters, and generate images in three \styles (Figure \ref{fig:image_ex}). In the \textit{color} style, the included characters are shown in red, while the excluded are black. In the \textit{left-right} style, a dashed line partitions the image into two halves. The included characters are on the right side and the excluded on the left. In the \textit{circle} style, the characters are arranged along the circumference of a circle: the included characters are inside and the excluded outside.

Exploring three ways of visualizing the partition allows us to measure how sensitive VLMs are to visual properties of the images. If a model performs well on one style but not the others, it would indicate that the model understands the pattern, but fails to apply it to certain kinds of images.

\subsection{Benchmark Construction}

We present \benchmark as a multiple choice Visual Question Answering task. Given an image and a multiple choice question with four pattern options, the goal is to pick the pattern that explains which characters are included and excluded. We randomly sample the three other answer options and manually verify that they are not correct patterns for the image, replacing them if so.

\begin{figure*}[t]
    \centering
    \includegraphics[width=\textwidth]{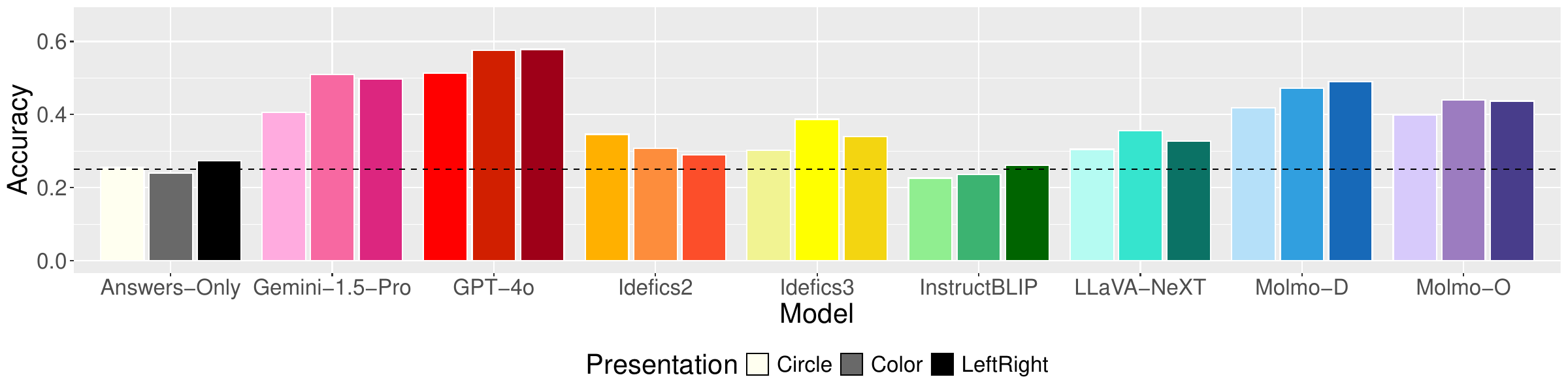}
    \caption{Model results by \style. Random baseline is plotted as a dashed line.}
    \label{fig:style}
\end{figure*}

\section{Experiments}\label{sec:experiments}

We use \benchmark to evaluate eight VLMs. All experiments are run with greedy decoding on a single A100 GPU. We evaluate eight models,\footnote{We also experimented with Kosmos-2~\cite{peng2024grounding} and OpenFlamingo~\cite{awadalla2023openflamingo}, but found that they did not understand the multiple choice task (Appendix \ref{app:other_models}).} both open and closed-source state-of-the-art VLMs: 

\begin{itemize}
\item \textbf{\gpt}~\cite{achiam2023gpt} is a closed-source multimodal chat model. 
\item \textbf{\gemini Pro}~\cite{reid2024gemini} is a 1.5T closed-source multimodal chat model. 
\item \textbf{\idefics}~\cite{laurenccon2024idefics2} is an 8B open-source multimodal chat model. 
\item \textbf{\ideficsN}~ is an improved version of \idefics.
\item \textbf{\llava}~\cite{liu2024llavanext} is an open-source multimodal LLaVA model with Mistral-7B-Instruct-v0.2.  
\item \textbf{\blip}~\cite{dai2024instructblip} is an open-source vision-language model that uses BLIP-2 with Vicuna-7b. 
\item \textbf{Molmo}~\cite{molmo2024} is a family of open-source multimodal language models; we benchmark Molmo-7B-O and Molmo-7B-D.
\end{itemize}

\subsection{Multiple Choice Experiments}

We benchmark all models using the multiple choice format of \benchmark. We use a chat template for chat models, with a system prompt explaining the task. For non-chat models, we suffix the instructions to the prompt.

\paragraph{Answers-Only Baseline}

In multiple choice tasks, information about the correct answer may be gleaned from the answer options. We measure this in a pre-experiment: we prompt \gpt with the answer choices alone, excluding the question and using an empty image. Accuracy is close to random. We call this the \textit{Answers-Only} baseline.

\paragraph{Zero-Shot Prompting}

We conduct zero-shot experiments by presenting the model with question and answer options, followed by the image.

\paragraph{Few-Shot Prompting}

For models that support few-shot prompting, we explore  prompting with one, three, and five examples. We format the few-shot prompt as a multi-turn message series. To avoid biasing the model, we use as examples patterns that do not reoccur in other writing systems. For example, we use a rule from the Zanabazar Square script, \textit{Characters that contain at least one enclosed rectangle}, rather than the commonly-occurring rule \textit{Characters with enclosed loop or circle}.

\paragraph{Chain-of-Thought Reasoning}

We also explored chain-of-thought reasoning in two models: \gpt and \gemini; details are in Appendix \ref{app:cot}.

\begin{table}[t]
\centering
\begin{tabular}{|l|l|l|l|l|l|}
\hline
\textbf{Model} & \textbf{0-shot} & \textbf{1} & \textbf{3} & \textbf{5}\\
\hline
\gemini& 47.1 & 52.0 &\textbf{53.7} &53.4\\
\gemini CoT&46.6&-&39.3&-\\
\hline
\gpt & \textbf{55.6} & 53.4 & 55.4 & 52.9\\
\gpt CoT&50.0&-&49.1&-\\
\hline
\idefics & \textbf{31.5} & 30.1 & 31.2 & 30.0 \\
\hline
\ideficsN & \textbf{34.3} & 33.5 & 32.7 & 33.3 \\
\hline
\blip & 24.1 & - & - & - \\\hline
\llava & \textbf{32.9} & 27.0 & 28.0 & 26.8\\
\hline
\molmoo & 42.6 & - & - & - \\
\molmod & \textbf{46.0} & - & - & - \\\hline
\end{tabular}
\caption{Zero-shot and few-shot accuracy by model.}
\label{tab:performance}
\end{table}

\subsection{Free Response Pattern Generation}

We also conduct a pattern description generation experiment with the two best-performing models. We prompt models with three image/pattern pairs and ask them to produce a rule for a fourth image. %Both models were given a system prompt explaining the task.

\section{Results}\label{sec:results}

We use \benchmark to measure abstract pattern understanding in eight VLMs. We test statistical reliability at a significance threshold of $\alpha$=0.05 using mixed-effects logistic regression models for each VLM (see Appendix \ref{app:mem}).

\subsection{Zero-shot Results}

We benchmark eight VLMs on \benchmark using the zero-shot multiple choice format. The benchmark is challenging for all models: even the best model, \gpt, selects the correct pattern little more than half of the time, showing that there is significant room for improvement in the abstract pattern recognition abilities of VLMs.

Table \ref{tab:performance} shows the zero-shot multiple choice results. \gpt is the best-performing model, followed by \gemini. \molmod is competitive with \gemini. The open-source \idefics, \ideficsN, and \llava models perform less well, but still better than chance. \blip does not outperform the answers-only baseline. 

Figure \ref{fig:style} displays the zero-shot results by \style. For most models, the circular partition is the most challenging, with a particularly substantial performance gap for \gemini ($\sim$10\%; $p$=0.006). The difference is smaller but statistically reliable for \gpt ($\sim$6\%; $p$=0.03). For \idefics, the left-right style is slightly more challenging.

\begin{table}[t]
\centering
\begin{tabular}{|l|l|l|l|l|l|}
\hline
\textbf{Model} & \textbf{Circle} & \textbf{Color} & \textbf{L/R} & \textbf{Total} \\
\hline
\gpt & 12.3 & 18.1 & 17.2 & 15.9 \\
\gemini &  6.9 &  8.3 &  9.9 &  8.4 \\
\hline
\end{tabular}
\caption{Validity of generated patterns by visual styles.}
\label{tab:freeresponse_performance}
\end{table}

\subsection{Does Few-shot Prompting Help?}

Table \ref{tab:performance} shows the few-shot multiple choice results for models that support it.\footnote{\molmo and \blip do not support few-shot.} The impact of few-shot prompting is inconsistent: \gpt shows no improvement and performance drops for \idefics, \ideficsN, and \llava. However, few-shot prompting benefits \gemini, bringing its performance close to \gpt. The impact was greater for \gemini's performance on the circle and left-right \styles. Surprisingly, more examples do not always help: \gemini does best with three.  

The inconsistent gains that we observe for few-shot prompting may be surprising, but similar findings have been reported for other VLM benchmarks, such as \citet{wu2024surprising}. \citet{wu2024surprising} find that Gemini benefits from few-shot prompting, while Idefics and GPT-4V do not. Given the converging evidence from different benchmarks, we think that this indicates an actual limitation of existing VLMs.

\subsection{Does Chain-of-Thought Reasoning Help?}

We also tested whether chain-of-thought reasoning improved the performance of the best models, \gemini and \gpt. We find that it actually hurts performance (Table \ref{tab:performance}), likely because, as we discuss in Section \ref{sec:reasoning}, models often produce incorrect reasoning. 

\subsection{Can Models Describe Patterns?}\label{res:free}

The multiple choice version of \benchmark explores whether VLMs can \textit{understand and assess} patterns described in natural language. We can also use \benchmark to assess whether models can \textit{describe} patterns. 

We explore the pattern description ability of the two best-performing VLMs, \gpt and \gemini, in a generation task. We prompt models with three image/pattern pairs and ask them to produce a rule for a fourth image. The authors independently assessed the validity of the patterns, with inter-annotator agreement of $\kappa$=0.89 for \gpt and 0.88 for \gemini.

Table \ref{tab:freeresponse_performance} presents the results, averaged over annotators. We find that \gpt is able to generate valid patterns for just 15.9\% of the images, and \gemini for 8.4\%. We also find that models struggle with the same \styles in pattern generation as in the multiple-choice task. 

Manual inspection reveals that models typically generate patterns similar to human ones: both models generated rules related to enclosed spaces, symmetry, and line type, which are common in \benchmark. Despite generating human-like rules, models struggle to apply them. This points to issues in pattern recognition or in text/image mapping.

\begin{figure}[t]
    \includegraphics[width=\columnwidth]{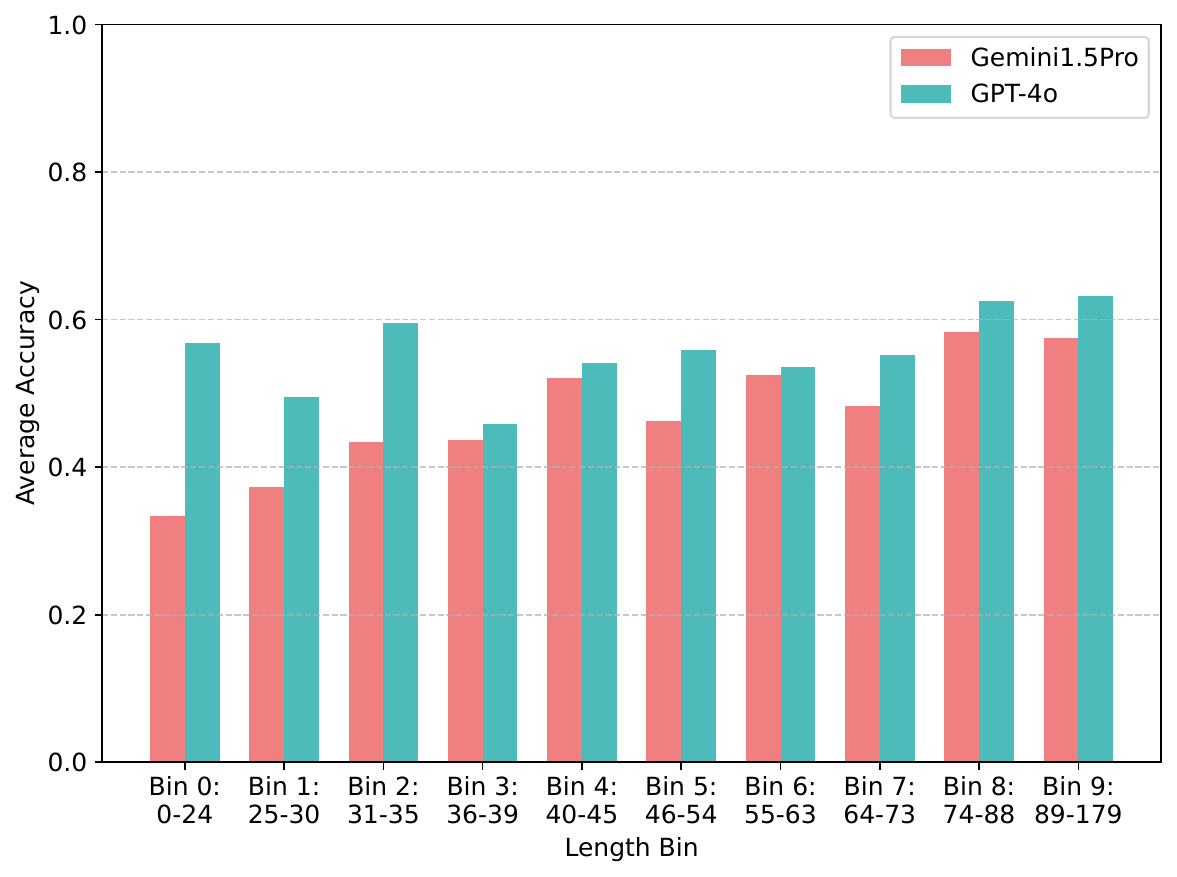}
    \caption{Accuracy by pattern length.}
    \label{fig:length}
\end{figure}

\section{Exploring Model Errors}

We have shown that \benchmark is a challenging benchmark for VLMs. The rich human-written rules and multiple \styles of \benchmark allow us to explore how various aspects of the task impact performance. In this section, we highlight key challenges that VLMs face in abstract pattern recognition, focusing on the two best-performing models, \gemini and \gpt. 

\subsection{Are Complex Descriptions Harder?}

One potential source of error is in understanding textual descriptions of patterns. A model may be able to distinguish visual properties of included versus excluded characters, but fail to understand how they are described in text. To explore this possibility, we investigate the \textit{complexity} of the pattern descriptions. Are simpler descriptions easier?

\begin{figure}[t]
    \includegraphics[width=\columnwidth]{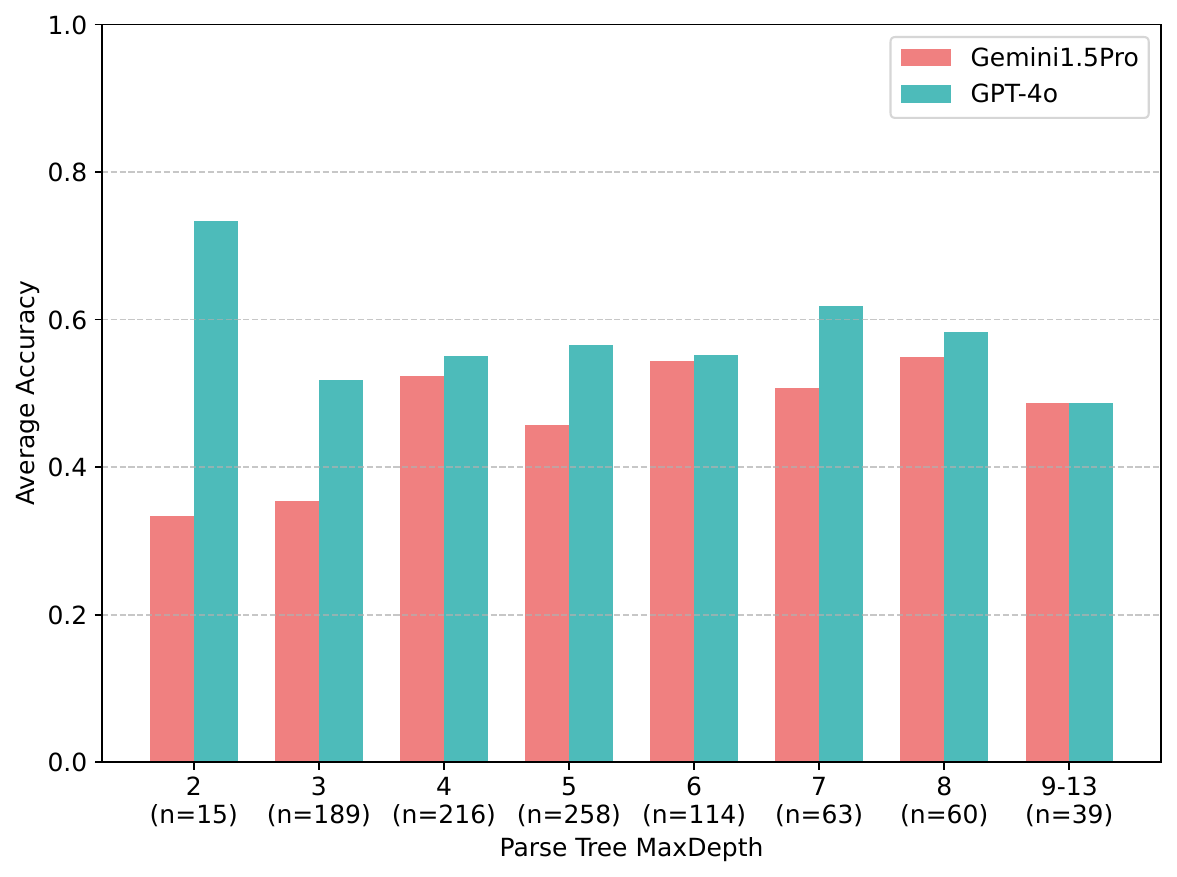}
    \caption{Accuracy by pattern parse tree depth.}
    \label{fig:depth}
\end{figure}

Figure \ref{fig:length} plots the performance of the two best-performing models against the number of characters in each rule as a proxy for complexity. Contrary to expectations, we observe that accuracy increases slightly with description length, suggesting that longer patterns are not more challenging. Mixed-effects models reveal a weak but statistically reliable correlation for both \gemini ($\widehat{\beta}$=0.003; $p$=0.001) and \gpt ($\widehat{\beta}$=0.003; $p$=0.002).

Another possibility is that models struggle with compositional language in the patterns. We quantify the syntactic complexity of each rule as the maximum depth of their dependency parse tree. Figure \ref{fig:depth} shows the performance of \gemini and \gpt by parse tree depth. We see no strong correlation in performance for either model, though \gpt performs best on very simple rules. 

\begin{figure}[t]
    \includegraphics[width=\columnwidth]{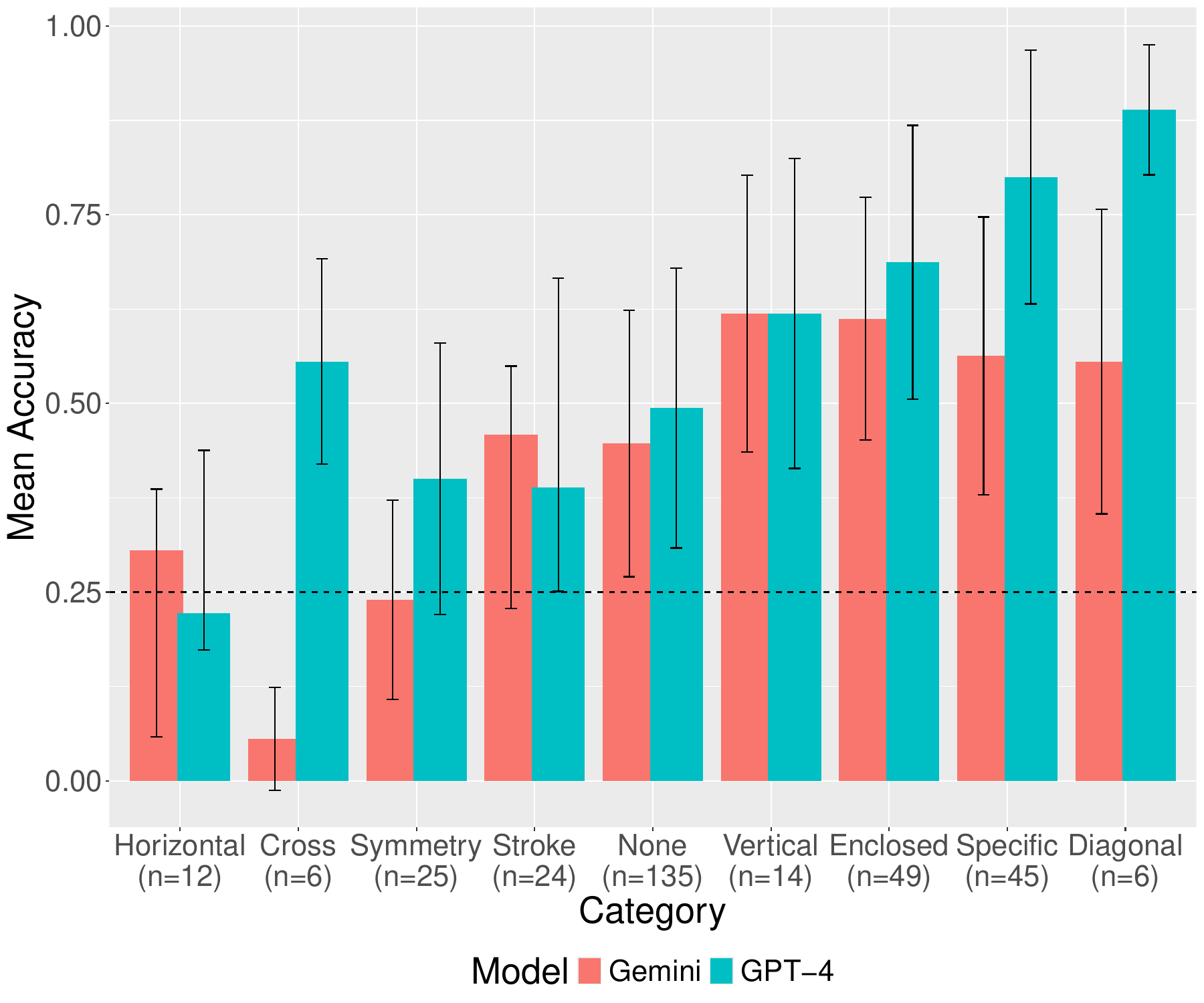}
    \caption{Accuracy by cross-linguistic pattern}
    \label{fig:patterns}
\end{figure}

\subsection{Do Models Understand Cross-linguistically Common Patterns?}\label{sec:cats}

\begin{figure*}[t]
    \includegraphics[width=\textwidth]{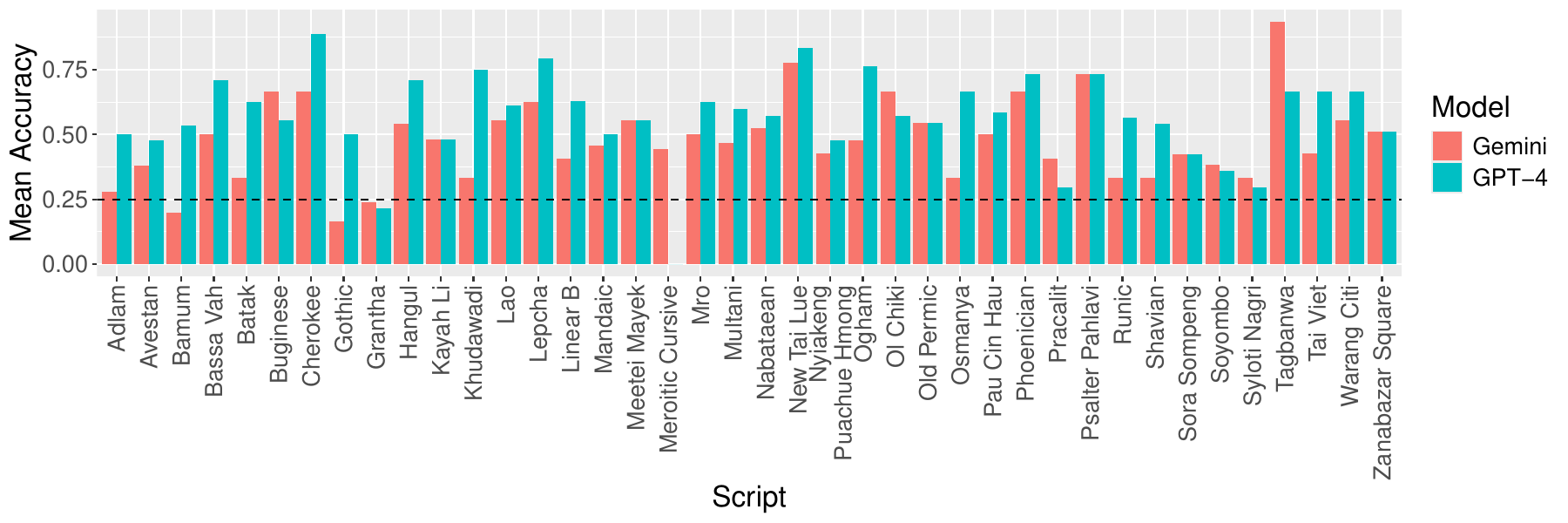}
    \caption{Accuracy by writing system.}
    \label{fig:langs}
\end{figure*}

The \textit{Glyph} project aims to explore cross-linguistic universals in visual properties of human writing systems. Towards this end, they have identified categories of rules that players commonly propose across languages. These categories represent visual properties of writing systems that are salient for humans. Are they also visually salient to VLMs?

Figure \ref{fig:patterns} plots performance by rule category for the two best-performing models. We observe significant variation in accuracy by rule category, showing that many rule categories commonly identified by humans pose challenges for models. 

The easiest kind of patterns for \gpt involve specific character mentions: for instance, \textit{Resemble the letter E}. Compared to rules that did not belong to a category, accuracy was higher ($\widehat{\beta}$=0.37; $p$\textless0.001). For \gemini, the easiest pattern was vertical lines ($\widehat{\beta}$=0.28; $p$=0.009). 

One surprising finding is the large difference for both models between vertical and horizontal lines. Accuracy was low in the horizontal category for both models, though only significantly different for \gpt (\gemini: $\widehat{\beta}$=-0.08; $p$=0.45; \gpt: $\widehat{\beta}$=-0.23; $p$=0.03). It is not clear why this line orientation is so challenging. Our results show that symmetry is also a difficult property.

It is less surprising that models struggle with the Stroke category, which describes how many pen strokes it would take to draw the character. These rules are grounded in the experience of drawing letters by hand, which VLMs lack. 

\subsection{What Patterns Do Models Produce Well?}

Our free response experiment allows us to examine what kinds of patterns \gemini and \gpt are most successful at generating. As discussed in Section \ref{res:free}, models often generate rules using similar language as humans even though their rules are usually incorrect for the target image. 

We categorize the valid patterns generated by each model according to the cross-linguistically common categories from Section \ref{sec:cats} (full results in Appendix \ref{app:free}). We find that the most common correctly generated patterns are ones that refer to enclosed spaces or specific characters. \gemini is more successful at identifying symmetry than \gpt. 

Qualitatively, we note that models seem to struggle especially to generate relative spatial terms. There are many generated patterns that would be correct if ``horizontal'' and ``vertical'' or ``left'' and ``right'' were swapped. This suggests that relative spatial language remains a challenge. Alternatively, these struggles could spring from a loss of spatial information in image-processing.

\subsection{Which Writing Systems Are Challenging?}

We also explore whether certain writing systems are more challenging for VLMs. Figure \ref{fig:langs} shows the performance of \gemini and \gpt by writing system. We see that some scripts are harder for them than others: \gpt struggles with Grantha ($\widehat{\beta}$=-0.4; $p$=0.02) and Meroitic Cursive ($\widehat{\beta}$=-0.5; $p$=0.03). This suggests that the visual properties of these writing systems are less salient for VLMs. 

\subsection{How Does Reasoning Go Awry?}\label{sec:reasoning}

When VLMs make mistakes, they can stem from issues in natural language understanding, image processing, or the vision-language mapping. To explore sources of error in \benchmark decisions, we prompt \gemini and \gpt to explain their reasoning for the multiple-choice task and manually inspect the results. We find issues stemming from multiple aspects of the task:

\paragraph{Errors in image-processing} In one case, \gpt correctly identifies the four excluded characters when presented in the color style. However, in the circle style, \gpt lists all 22 characters in the script as falling inside the circle, indicating an error in recognizing the circle itself (see Appendix \ref{app:errors}). This is similar to the image-processing errors in \citet{rahmanzadehgervi2024vision}.

\begin{figure}[t]
    \fbox{\includegraphics[width=0.95\linewidth]{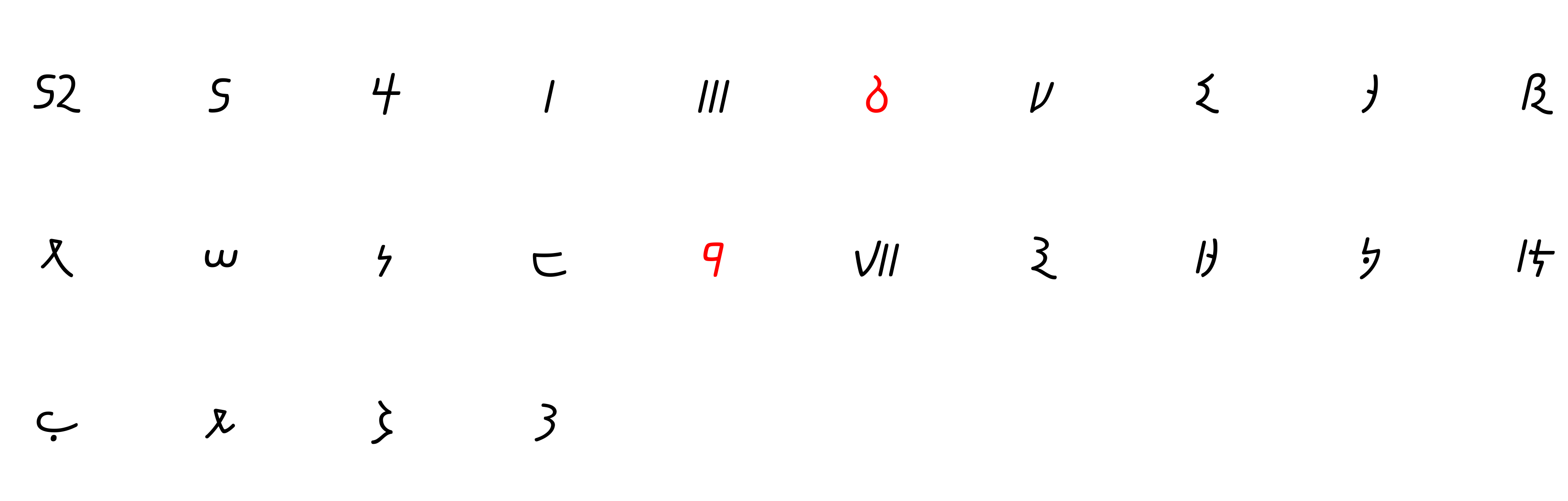}}\\

\textbf{Target Pattern}: \textit{Characters that have exactly one closed end.}

    \textbf{\gemini}: \textit{The characters `S', `2', `C', and `3' have exactly one closed end, but they are not colored red.}

    \textbf{\gpt}: \textit{The red characters do not have exactly one closed end. They have loops, which means they have more than one closed end.}
    
    \caption{Ex. natural language understanding failure}
    \label{fig:text-mistake}
\end{figure}

\paragraph{Errors in natural language understanding} In other cases, models fail to understand the language that human pattern writers use. For the example in Figure \ref{fig:text-mistake}, \gpt is able to map the included characters to their Unicode symbols and identify that these characters contain loops. However, \gpt rejects the correct option because it does not understand ``closed end'' as the human writer intended. \gemini also misunderstands this description and incorrectly identifies it as applying to multiple characters in the excluded set.

\paragraph{Text/image mapping issues}

\begin{figure}[t]
    \fbox{\includegraphics[width=0.95\linewidth]{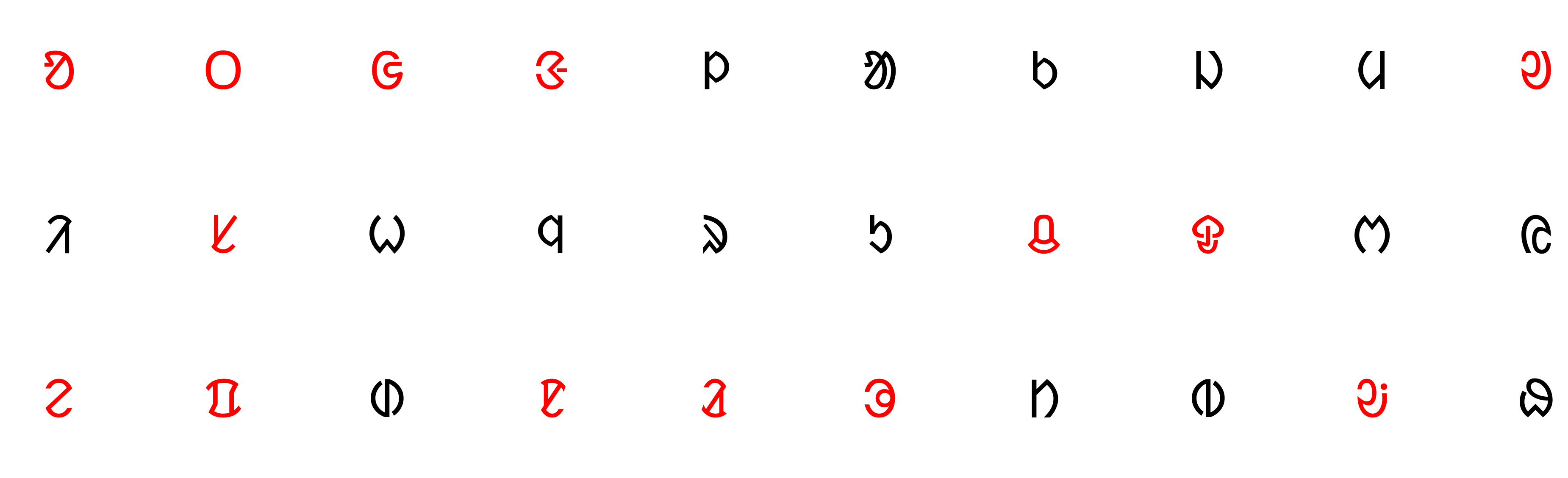}}\\
    
    \textbf{Target Pattern}: \textit{Characters that have a curved bottom.}
    
    \textbf{\gemini}: \textit{While some red characters have curved bottoms, others do not. For example, the red character "O" does not have a curved bottom.}
    
    \textbf{\gpt}: \textit{Not all red characters have a curved bottom. For example, the character in second row, second column from the left (which looks like an upside-down "L") does not have a curved bottom.}
    
    \caption{Ex. incorrect character property reasoning}
    \label{fig:prop-mistake}
\end{figure}

Models sometimes fail to map the properties described in natural language to the character images. For the curved bottom pattern in Figure \ref{fig:prop-mistake}, \gemini correctly picks out a red character that is similar to ``O'', but does not seem to understand the visual properties of the character or its Latin equivalent. \gpt makes a similar mistake for the same pattern, incorrectly deciding that the ``L''-like character lacks a curved bottom but does have ``enclosed spaces or loops" (one of the distractor options). Here, the models either fail to understand the property description, or to map it to the visual properties of the characters.

\paragraph{Reasoning errors} Other errors arise from failures to reason about the excluded characters. For instance, \gpt chooses the pattern \textit{Characters that look like Latin capital letters} for the item in Figure \ref{fig:image_ex}. Its explanation (\textit{The characters on the right side, ``U” and ``W”, look like Latin capital letters}) correctly judges that the included characters fit the pattern, but fails consider that excluded characters do too. 

Thus, \benchmark shows that VLM errors arise across the image-processing, text-image mapping, and reasoning aspects of the abstract pattern recognition task.

\section{Conclusion}

We present \benchmark, an abstract pattern recognition benchmark consisting of 954 items in three visual styles. \benchmark challenges models to understand visually-presented patterns paired with human descriptions, revealing significant gaps in VLM capabilities. We find that even the best performing models, \gemini and \gpt, solve around half of \benchmark problems. 

\benchmark is based on human-written descriptions of visual properties of writing systems, facilitating the exploration of multiple task aspects, from pattern length, complexity, and visual presentation style, to differences across cross-linguistically common pattern types and human writing systems. Our detailed analysis reveals errors at multiple levels of processing, including image-processing, natural language understanding, and reasoning. We hope that the \benchmark benchmark will guide future work on improving VLMs by providing a challenging, targeted and diverse evaluation set.

\section*{Limitations}

Our work reveals weaknesses in abstract pattern recognition for VLMs. \benchmark moves closer to Bongard's original task by requiring models to understand natural language pattern descriptions; however, evaluating the ability to describe patterns is challenging due to the lack of automated scoring. Automating this process is unlikely to succeed in the near-term: since our benchmark items are challenging for state-of-the-art VLMs, using LLMs to score the rules would not be reliable.

Our few-shot evaluation of \idefics was limited by GPU memory constraints. We also excluded two VLMs due to inability to understand the task format. It is possible that they could perform well in other abstract pattern recognition tasks, but ideally, models would be robust to task format.

\section*{Ethics Statement}

Our work does not involve any data collection from human subjects. We rely on publicly-available human data from the Glyph project, which was originally collected under the oversight of the Ethikkommission affiliated with Universitätsklinikum, ensuring adherence to ethical guidelines and standards. 

Our findings have implications for the visual salience of character sets across human writing systems, and we hope that they complement ongoing cognitive scientific efforts to understand the visual properties of human writing systems. Although our results show that some writing systems are more challenging for VLMs, this should not be misinterpreted to suggest that some writing systems are better or more efficient than others.

\section*{Acknowledgements}

We thank the ARR reviewers for their thoughtful feedback. This work is partially supported by the U.S. National Science Foundation (SES-2326174). 

\bibliography{anthology,custom}

\clearpage

\appendix

\section{Checklist}

A2 Potential Risks: 
Did you discuss any potential risks of your work? [Yes/No/NA]
\textbf{No}

A2 Elaboration:
For yes, provide a section number. For no, justify why not.
\textbf{The risks of this work lie in the original dataset creation and collection, as described by \citet{glyph}.}

B Use Or Create Scientific Artifacts:
Did you use or create scientific artifacts? [Yes/No]
\textbf{Yes}

B1 Cite Creators Of Artifacts:
Did you cite the creators of artifacts you used? [Yes/No/NA]
\textbf{Yes}

B1 Elaboration:
For yes, provide a section number. For no, justify why not.
\textbf{Section \ref{sec:dataset}.}

B2 Discuss The License For Artifacts:
Did you discuss the license or terms for use and/or distribution of any artifacts? [Yes/No/NA]
\textbf{Yes}

B2 Elaboration:
For yes, provide a section number. For no, justify why not. \textbf{Section \ref{artifact-url-and-license}}

B3 Artifact Use Consistent With Intended Use:
Did you discuss if your use of existing artifact(s) was consistent with their intended use, provided that it was specified? For the artifacts you create, do you specify intended use and whether that is compatible with the original access conditions? [Yes/No/NA]
\textbf{Yes}

B3 Elaboration:
For yes, provide a section number. For no, justify why not.
\textbf{Ethics Statement}

B4 Data Contains Personally Identifying Info Or Offensive Content:
Did you discuss the steps taken to check whether the data that was collected/used contains any information that names or uniquely identifies individual people or offensive content, and the steps taken to protect/anonymize it? [Yes/No/NA]
\textbf{NA}

B4 Elaboration:
For yes, provide a section number. For no, justify why not.

B5 Documentation Of Artifacts:
Did you provide documentation of the artifacts, e.g., coverage of domains, languages, and linguistic phenomena, demographic groups represented, etc.? [Yes/No/NA]
\textbf{Yes}

B5 Elaboration:
For yes, provide a section number. For no, justify why not.
\textbf{Section \ref{app:mem}.}

B6 Statistics For Data:
Did you report relevant statistics like the number of examples, details of train/test/dev splits, etc. for the data that you used/created? [Yes/No/NA]
\textbf{Yes}

B6 Elaboration:
For yes, provide a section number. For no, justify why not.
\textbf{Section \ref{sec:dataset}}

C Computational Experiments:
Did you run computational experiments? [Yes/No/NA]
\textbf{Yes}

C1 Model Size And Budget:
Did you report the number of parameters in the models used, the total computational budget (e.g., GPU hours), and computing infrastructure used? [Yes/No/NA]
\textbf{Yes}

C1 Elaboration:
For yes, provide a section number. For no, justify why not.
\textbf{Section \ref{computing-resources}}

C2 Experimental Setup And Hyperparameters:
Did you discuss the experimental setup, including hyperparameter search and best-found hyperparameter values?
[Yes/No/NA]
\textbf{Yes}

C2 Elaboration:
For yes, provide a section number. For no, justify why not.
\textbf{Section \ref{sec:experiments}.}

C3 Descriptive Statistics:
Did you report descriptive statistics about your results (e.g., error bars around results, summary statistics from sets of experiments), and is it transparent whether you are reporting the max, mean, etc. or just a single run? [Yes/No/NA]
\textbf{Yes}

C3 Elaboration:
For yes, provide a section number. For no, justify why not.
\textbf{Sections \ref{sec:results} and \ref{app:mem}.}

C4 Parameters For Packages:
If you used existing packages (e.g., for preprocessing, for normalization, or for evaluation, such as NLTK, SpaCy, ROUGE, etc.), did you report the implementation, model, and parameter settings used? [Yes/No/NA]
\textbf{Yes}

C4 Elaboration:
For yes, provide a section number. For no, justify why not. \textbf{Section \ref{app:complexity}.}

D Human Subjects Including Annotators:
Did you use human annotators (e.g., crowdworkers) or research with human subjects? [Yes/No/NA]
\textbf{No}

D1 Instructions Given To Participants:
Did you report the full text of instructions given to participants, including e.g., screenshots, disclaimers of any risks to participants or annotators, etc.? [Yes/No/NA]
\textbf{NA}

D1 Elaboration:
For yes, provide a section number. For no, justify why not.

D2 Recruitment And Payment:
Did you report information about how you recruited (e.g., crowdsourcing platform, students) and paid participants, and discuss if such payment is adequate given the participants' demographic (e.g., country of residence)? [Yes/No/NA]
\textbf{NA}

D2 Elaboration:
For yes, provide a section number. For no, justify why not.

D3 Data Consent:
Did you discuss whether and how consent was obtained from people whose data you're using/curating (e.g., did your instructions explain how the data would be used)? [Yes/No/NA]
\textbf{NA}

D3 Elaboration:
For yes, provide a section number. For no, justify why not.

D4 Ethics Review Board Approval:
Was the data collection protocol approved (or determined exempt) by an ethics review board? [Yes/No/NA]
\textbf{NA}

D4 Elaboration:
For yes, provide a section number. For no, justify why not.

D5 Characteristics Of Annotators:
Did you report the basic demographic and geographic characteristics of the annotator population that is the source of the data? [Yes/No/NA]
\textbf{NA}

D5 Elaboration:
For yes, provide a section number. For no, justify why not.

E Ai Assistants In Research Or Writing:
Did you use AI assistants (e.g., ChatGPT, Copilot) in your research, coding, or writing? [Yes/No]
\textbf{Yes}

E1 Information About Use Of AI Assistants:
Did you include information about your use of AI assistants? [Yes/No/NA]
\textbf{Yes}

E1 Elaboration:
For yes, provide a section number. For no, justify why not.
\textbf{Section \ref{use-of-ai-assistants}.}

\clearpage

\section{Further Details of Dataset}

\subsection{Licensing}\label{artifact-url-and-license}

The full GlyphPattern dataset and support code are available at \url{https://github.com/Wellesley-EASEL-lab/GlyphPattern}. The benchmark is licensed for reuse under an MIT license. 

The dataset builds on data collected as part of the Glyph cognitive science project, which is publicly available at \url{https://osf.io/j9zhn}.

\subsection{Data Selection and Cleaning}

We use the 318 cleaned rules from \citet{glyph}'s Replicability Assessment datafile. We further standardize the wording of the rules by hand, such that all are grammatical following the prefix \textit{Characters that}. This is necessary in order to consistently produce grammatical prompts.

We eliminated two scripts included in Glyph, Afaka and Miao, due to difficulties in obtaining font libraries containing their full character lists.

\section{Further Details of Experiments}

\subsection{Model Selection}\label{app:other_models}

We use the following model versions:

\begin{itemize}
\item \gpt: gpt-4o-2024-05-13
\item \gemini: gemini-1.5-pro-001
\item \llava: llava-v1.6-mistral-7b-hf
\item \idefics: idefics2-8b
\item \ideficsN: Idefics3-8B-Llama3
\item \blip: instructblip-vicuna-7b
\end{itemize}

For \idefics, we used image scaling parameters that set the longest edge of 700 and a shortest edge of 378. This did not decrease performance compared to the default \idefics settings (Table \ref{tab:scaling}).

For \ideficsN, we used image scaling parameter that set the longest edge of 1092. This did not decrease performance compared to the default \ideficsN settings (Table \ref{tab:scaling}).

\begin{table}[H]
    \centering
    \begin{tabular}{|l|l|l|}\hline
         Model&Image Scaling&Accuracy (\%)\\\hline
         \idefics&base&31.45\\
         \idefics&scaled&31.45\\\hline
         \ideficsN&base&33.86\\
         \ideficsN&scaled&34.28\\\hline
    \end{tabular}
    \caption{Zeroshot accuracy for \idefics and \ideficsN with and without image scaling.}
    \label{tab:scaling}
\end{table}

We attempted to benchmark Kosmos-2~\cite{peng2024grounding} and OpenFlamingo~\cite{awadalla2023openflamingo}, but found that they struggled with the multiple choice task format. Both models tended to produce additional answer options rather than selecting an option from among the four choices. Adding the task explanation to the end did not address this issue; instead, Kosmos-2 generated a new explanation without an answer option, and OpenFlamingo began generating a new set of answers starting at A.

We also attempted to benchmark Llama3.2 Vision~\cite{llama3p2modelcardvision}, but found that it is safeguard to the extend of not providing valid response for our test. Instead, the model respond "I'm not able to provide that information".

\subsection{Computing Resources}\label{computing-resources}

Experiments were run with greedy decoding on a single A100 GPU. We estimate that the experiments reported in this paper consumed less than 400 GPU hours; however, we do not know how much GPU time was used to run the proprietary models \gemini and \gpt.

\begin{table}[t]
\centering
\begin{tabular}{|l|l|l|l|l|l|}
\hline
\textbf{Model} & \textbf{N} & \textbf{Circle} & \textbf{Color} & \textbf{L/R} & \textbf{Total}\\
\hline
Gemini& 0 & 40.6 & 51.0 & 49.7&47.1 \\

Gemini& 1 & 49.1 & 50.6 & 56.3&52.0 \\

Gemini& 3 & 50.3 & 52.2 & 58.5 &\textbf{53.7}\\

Gemini& 5 & 50.6 & 52.5 & 57.0 &53.4\\
\hline
\gpt& 0 & 51.3 & 57.6 & 57.9 &\textbf{55.6}\\

\gpt& 1 & 48.5 & 57.6 & 54.1 &53.4\\

\gpt& 3 & 49.7 & 57.6 & 59.1 &55.4\\

\gpt& 5 & 45.9 & 58.5 & 54.4 &52.9\\
\hline
\idefics & 0& 34.6 & 30.8 & 29.0 &\textbf{31.5}\\
\idefics & 1 & 33.0 & 30.8 & 26.4&30.1 \\
\idefics & 3 & 34.0 & 29.9 & 29.9&31.2 \\
\idefics & 5 & 33.0 & 30.5 & 26.4&30.0 \\
\hline

\ideficsN & 0 & 30.2 & 38.7 & 34.0 &\textbf{34.3}\\
\ideficsN & 1 & 31.5 & 32.7 & 36.5&33.5 \\
\ideficsN & 3 & 32.7 & 34.6 & 30.8&32.7 \\
\ideficsN & 5 & 33.3 & 35.0 & 31.8&33.3 \\
\hline

LLaVA & 0 & 30.5 & 35.5 & 32.7 &\textbf{32.9}\\

LLaVA & 1 & 28.3 & 24.8 & 28.0 &27.0\\

LLaVA & 3 & 29.6 & 27.0 & 27.4 &28.0\\

LLaVA & 5 & 26.4 & 26.4 & 27.7 &26.8\\
\hline
\end{tabular}
\caption{Few-shot accuracy by number of examples (N)}
\label{tab:fewshot_performance_style}
\end{table}

\subsection{Further Prompting Details}

We ran \idefics,  \ideficsN, \llava, and \blip using Hugging Face's AutoModelForVision2Seq API, following the format of their model card. 

\paragraph{System Prompts}

For multiple choice experiments, we query the model with system prompt: \textit{Reply with only the letter of the correct option.} The system prompt is provided to \gpt and \gemini through their APIs. 
We provide no system prompt for \idefics or \ideficsN. For \llava, and \blip zero-shot experiments, we concatenated the system prompt at the end of the question prompt for zero-shot experiments. We do not provide a system prompt for their few-shot experiments.

For few-shot free response experiments, we prompt \gpt and \gemini with the system instruction: \textit{Come up with a descriptions for all of the characters \{on the right\} side in the image. Your description should be true for all of the characters \{on the right side\} and none of the characters \{on the left side\}}, reworded by the image's \style.

\paragraph{Few-shot Prompting}

For \idefics, \ideficsN, \llava, and \gemini, the image is presented before the question, followed by the answer. For \gpt, the question is shown first, then the image and the answer. 

\begin{figure}[h]
    \fbox{%
        \begin{minipage}{\linewidth}
            {\centering
            \includegraphics[width=0.8\linewidth]{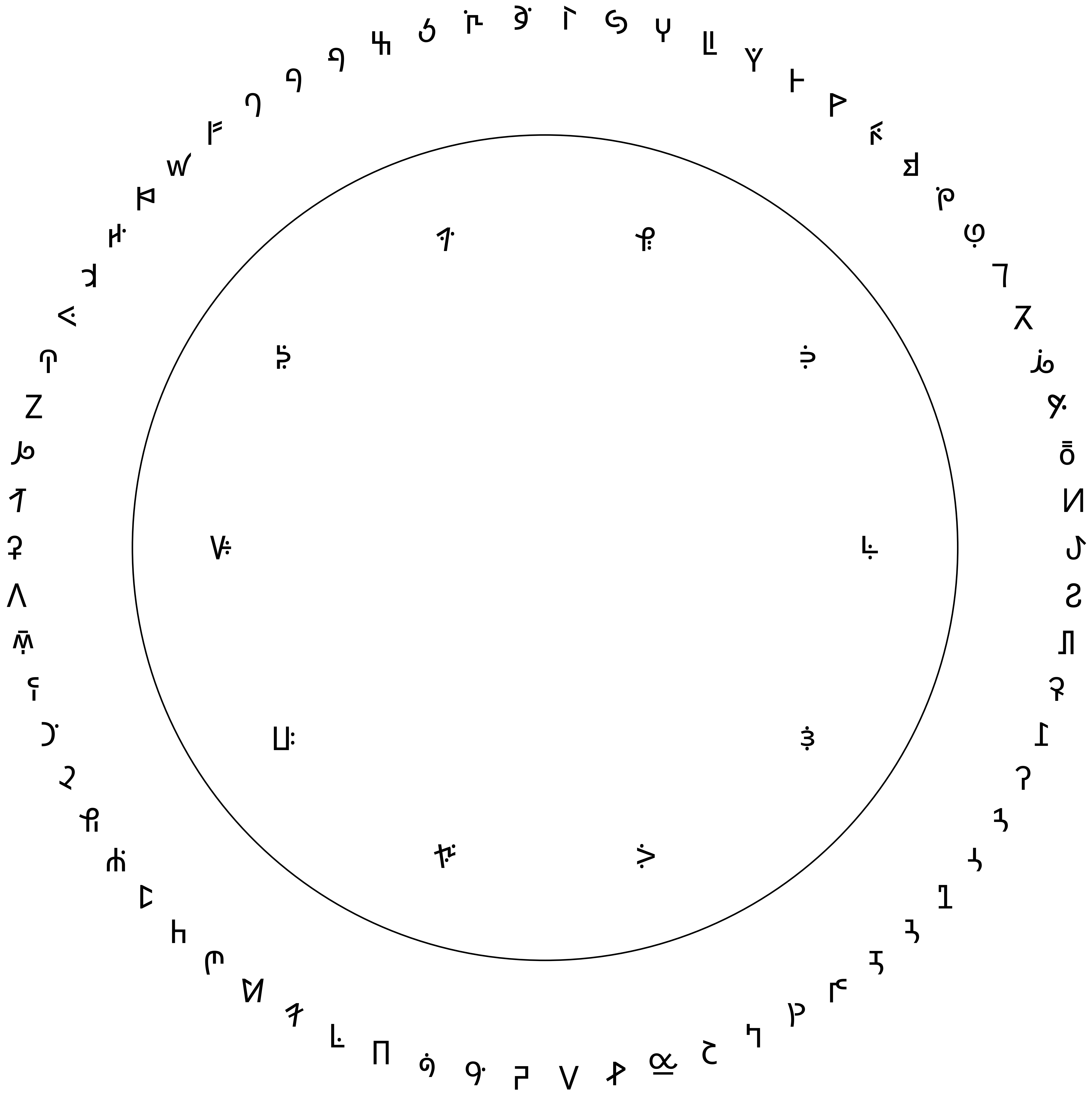}\\}
            \vspace{0.05\linewidth}
            {\raggedright\textit{Which characters are inside the circle in the image? A. Characters that contain a horizontal line and a descending curve from one end of the line. B. Characters that have a downward curve that resembles a bridge with two feet. C. Characters that have two round dots. D. Characters that contain diagonal lines and acute angles, but no flags.}\\
            \textit{MODEL:C.}\\}
            \vspace{0.05\linewidth}
            {\centering
            \includegraphics[width=0.8\linewidth]{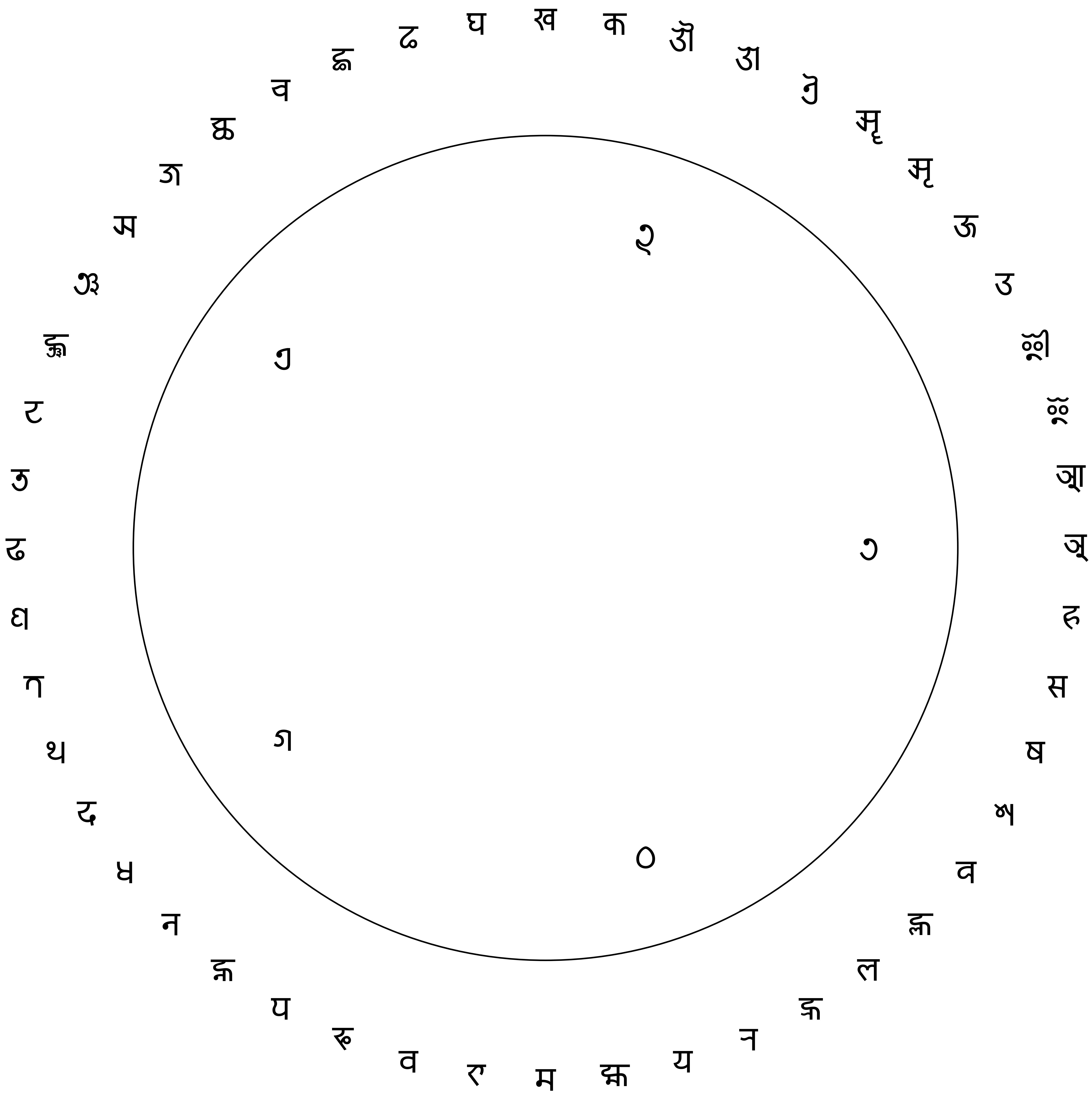}\\}
            \vspace{0.05\linewidth}
            \textit{Which characters are colored red in the image? A. Characters that contain a hook shape. B. Characters that start at the top left with a small or very small spiral-like approach. C. Characters that can be written with a single continuous stroke without retracing or lifting the pen. D. Characters that have vertical and horizontal symmetry.}\\
            \textit{MODEL}:\\
            \textbf{Groundtruth}:\textit{C.}\\
            \textbf{\gemini Response}:\textit{C.}\\
            \textbf{\gpt Response}:\textit{D.}
            
        \end{minipage}
    }
    \caption{Few-shot Multiple Choice Example}
    \label{fig:multiplechoiceex}
\end{figure}

Figure \ref{fig:multiplechoiceex} shows an example of few-shot multiple choice prompting with one sample question and answer pair. 

\begin{figure}[h]
    \fbox{%
        \begin{minipage}{\linewidth}{\centering\includegraphics[width=0.7\linewidth]{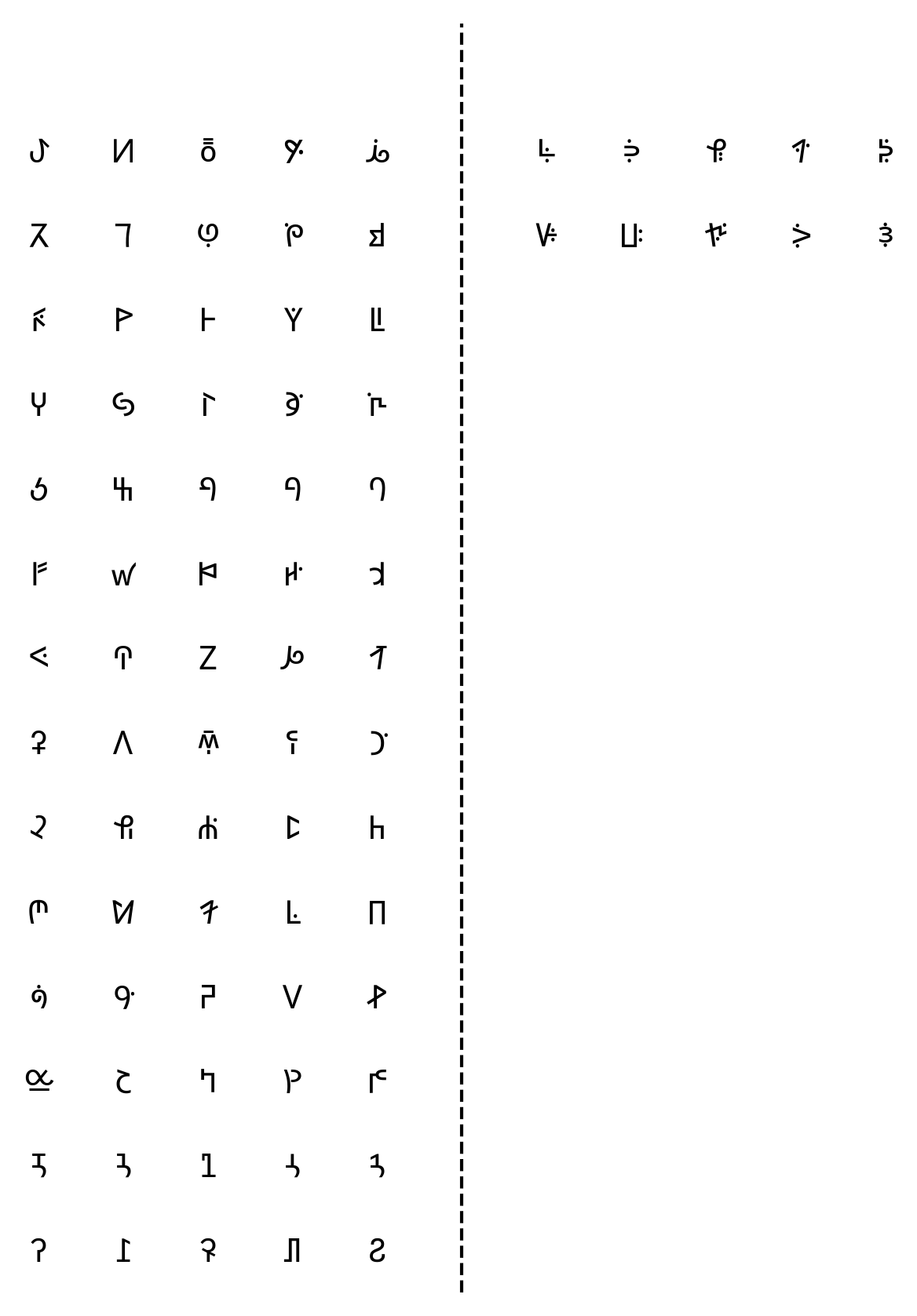}\\}
            \textit{Characters on the right in the image are characters that have two round dots.}\\
            {\centering\includegraphics[width=0.7\linewidth]{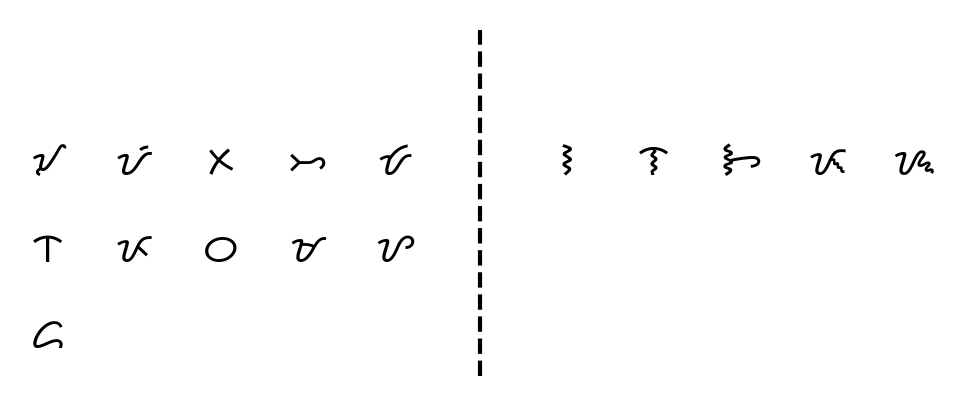}\\}
            \textit{Characters on the right in the image are characters that contain a wavy or zigzag line.}\\
            {\centering\includegraphics[width=0.7\linewidth]{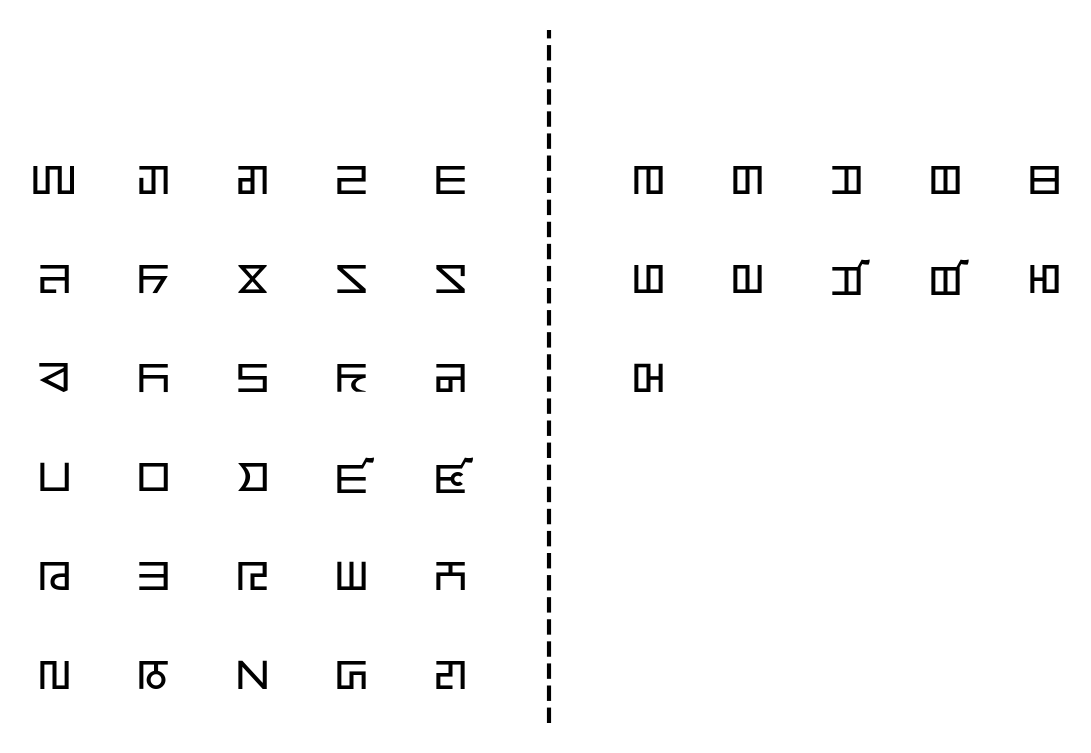}\\}
            \textit{Characters on the right in the image are characters that contain at least one enclosed rectangle.}\\
            {\centering\includegraphics[width=0.7\linewidth]{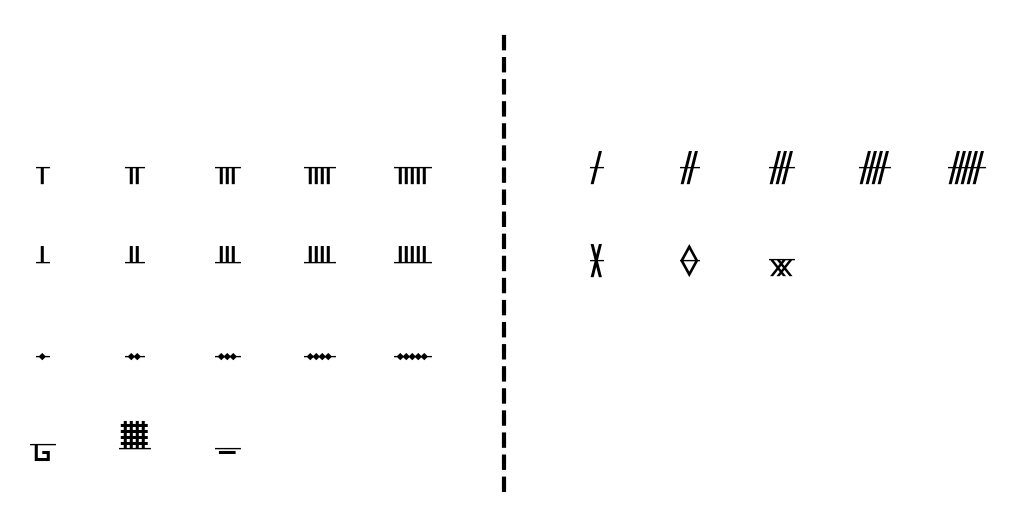}\\}
            \textbf{Groundtruth}:
            \textit{contain at least one diagonal line.}\\
            \textbf{\gemini Response}:
            \textit{Characters on the right side of the image are angled or diagonal.}\\
            \textbf{\gpt Response}:
            \textit{Characters on the right side in the image are characters that contain diagonal lines.}\\
            \end{minipage}
            }
    \caption{Few-shot Free Response Example}
    \label{fig:freeresponseex}
\end{figure}

Figure \ref{fig:freeresponseex} is an example of few-shot free response.

\subsubsection{Chain-of-Thought Reasoning Prompting}\label{app:cot}

\begin{figure}[h]
\fbox{%
        \begin{minipage}{\linewidth}
            
        {\centering\includegraphics[width=0.7\linewidth]{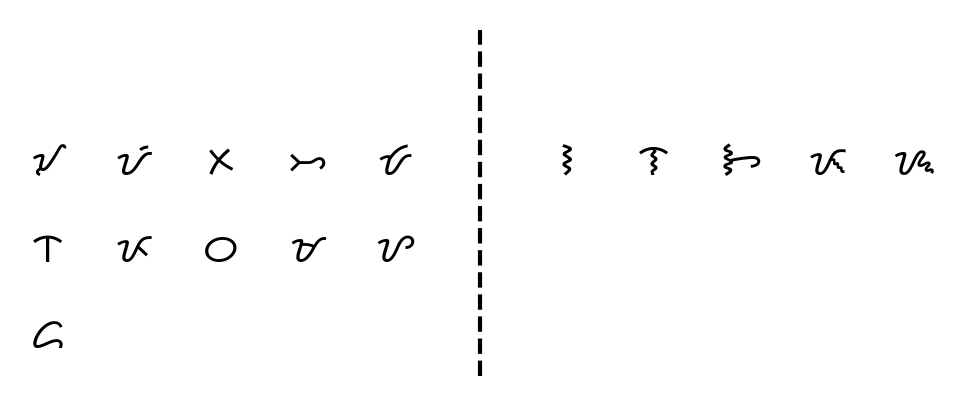}\\}
            
        210 Left/Right: Let's think this through step-by-step. All of the characters on the right do contain wavy or zigzag lines and none of the characters on the left do, so option A is a correct description. Some of the shapes on the right don't have loops, so option B is incorrect. Some of the shapes on the right do not contain a U or a hook, so option C is incorrect. Some of the shapes on the right do not have a loop at the bottom, so option D is also incorrect. Since option A is the only pattern that describes all characters on the right and none of the characters on the left, it must be the right answer.

        {\centering\includegraphics[width=0.7\linewidth]{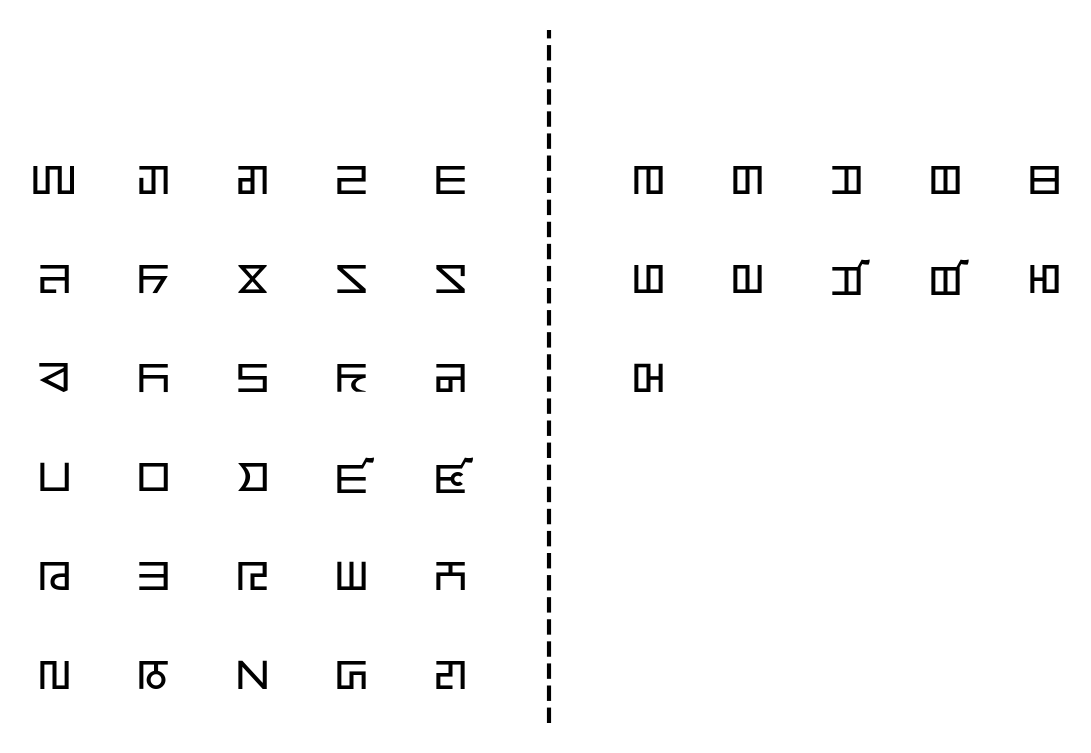}\\}

        298 Left/Right: Let's think this through step-by-step. Almost all of the characters, both on the right and on the left, have a vertical straight line, so option A is incorrect. All of the characters on the right contain at least one enclosed rectangle, so option B is possible. Also, none of the ones on the left have an enclosed rectangle, so option B is valid. Many of the characters on the right cannot be drawn with a single stroke, so option C is incorrect. Similarly, some of the characters on the right do not have an open area facing up, so option D cannot be the answer. Since option B is the only pattern that describes all characters on the right and none of the characters on the left, it must be the right answer.
    \end{minipage}
    }
    \caption{Chain-of-Thought Reasoning Examples for Few-shot Prompting}\label{ex:cot}
\end{figure}

We ran zero-shot and three-shot experiments with chain-of-thought prompting. Two example reasoning chains used for chain-of-thought few-shot prompting are shown in Figure \ref{ex:cot}. 

\clearpage

\section{Further Analysis}

\subsection{Pattern Complexity}\label{app:complexity}

We compute the syntactic complexity of a pattern by taking the maximum depth of its dependency parse tree. We parse the pattern descriptions using SpaCy (``en\_core\_web\_lg'' model).

\subsection{Error Analysis}\label{app:errors}
\begin{figure*}[t]
\fbox{\begin{minipage}[t]{\dimexpr0.9\textwidth-.5\columnsep}
    \textbf{Target Pattern}: \textit{Characters that have a rough triangle pointing lower-left}\\
    \fbox{\includegraphics[width=0.45\linewidth]{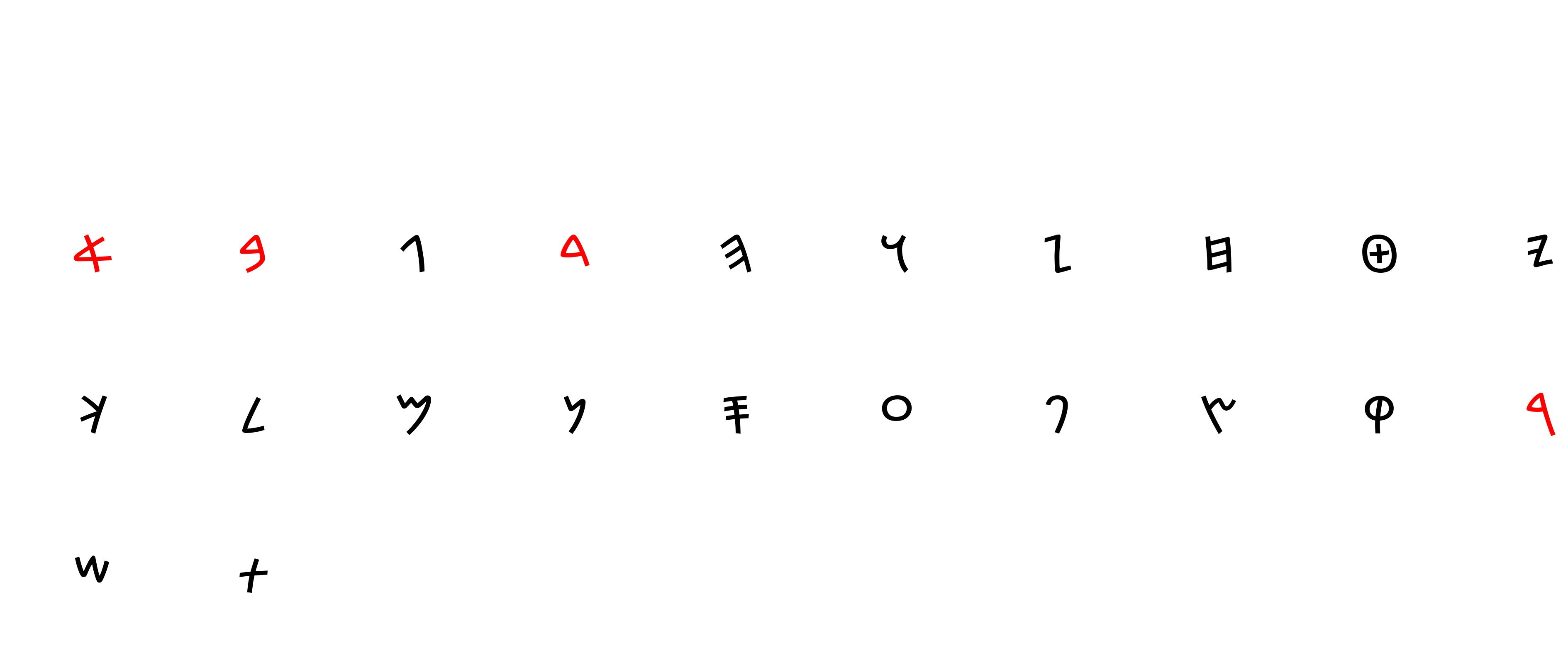}}\\
    
    \textbf{\gpt Color Explanation}: \textit{The red characters in the image all have a shape that resembles a triangle pointing towards the lower-left direction.}\\

    \fbox{\includegraphics[width=0.4\linewidth]{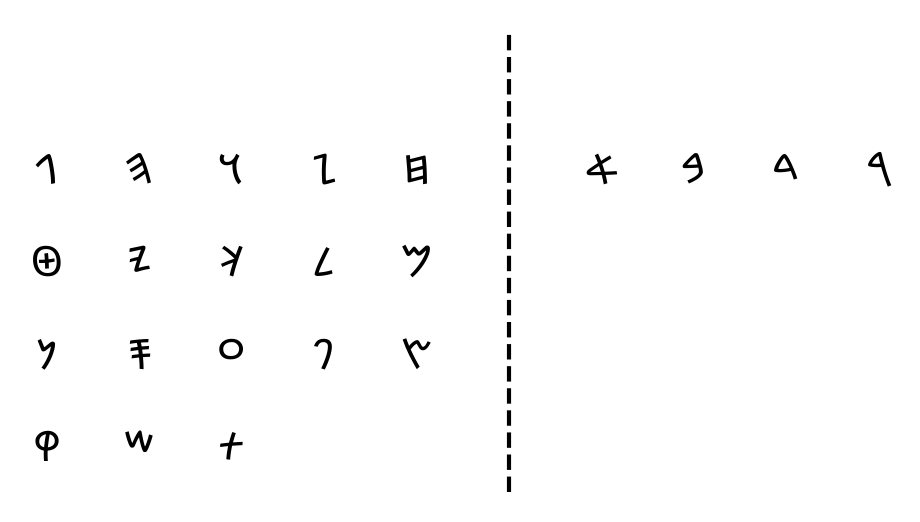}}\\
    
    \textbf{\gpt Leftright Explanation}: \textit{All the characters on the right side have a rough triangle pointing lower-left. The characters resemble "4", "9", and "A", which all have a triangular shape pointing towards the lower-left.}\\
    
    \fbox{\includegraphics[width=0.45\linewidth]{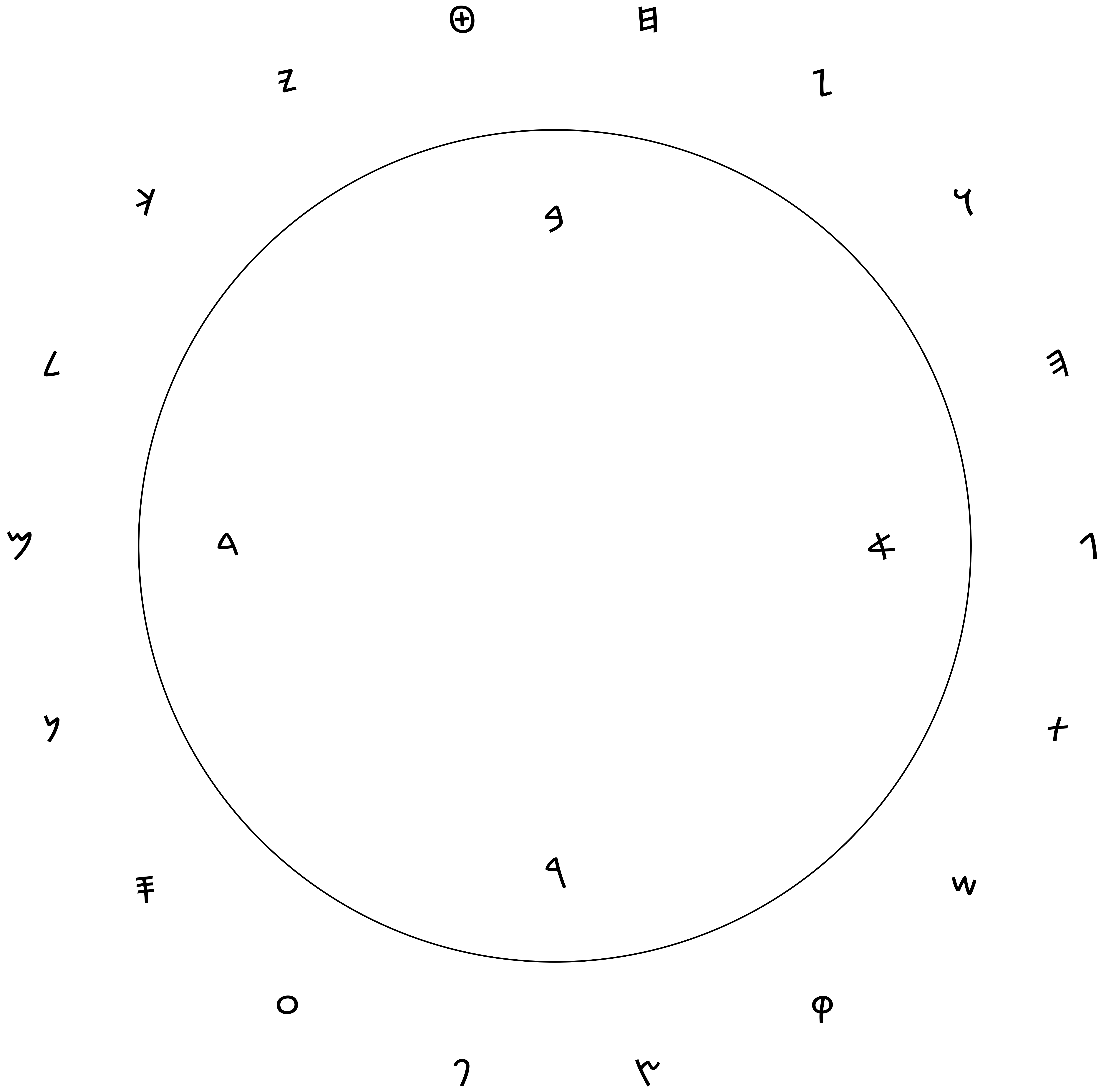}}\\
    
    \textbf{\gpt Circle Explanation}: \textit{The characters inside the circle are\textphnc{A,B,G,D,H,W,Z,Ch,T,Y,K,L,M,N,S,P,C,Q,R,S,Th}. They do not all have a rough triangle pointing lower-left.}
    \end{minipage}}
    \caption{Example of image-processing error}
    \label{fig:visual-mistake}
\end{figure*}

Figure \ref{fig:visual-mistake} illustrates a case where the circular \style poses a challenge for the models. \gpt performed correct visual recognition of the excluded character sets on both the color and leftright \style, but failed on the circle \style.

\subsection{In-Depth Few-shot Results}

Table \ref{tab:fewshot_performance_style} shows few-shot results by model and \style.

\subsection{In-Depth Chain-of-Thought Results}

Table \ref{tab:cot_performance_style} shows the chain-of-thought reasoning results by model and \style.

\begin{table}[h]
    \centering
    \begin{tabular}{|l|l|l|l|l|l|}\hline
         Model&N&Circle&Color&L/R&Total\\\hline
         \gpt&0&45.0&54.7&50.3&50.0\\
         \gpt&3&40.3&51.9&55.0&49.1\\
         \gemini&0&39.3&50.6&50.0&46.6\\
         \gemini&3&35.2&39.0&43.7&39.3\\\hline
    \end{tabular}
    \caption{Chain-of-Thought Prompting Results by Model}
    \label{tab:cot_performance_style}
\end{table}

\subsection{Free Response Task}\label{app:free}

\begin{figure}[H]
    \centering
    \begin{minipage}[b]{0.45\textwidth}
        \centering
        \includegraphics[width=\linewidth]{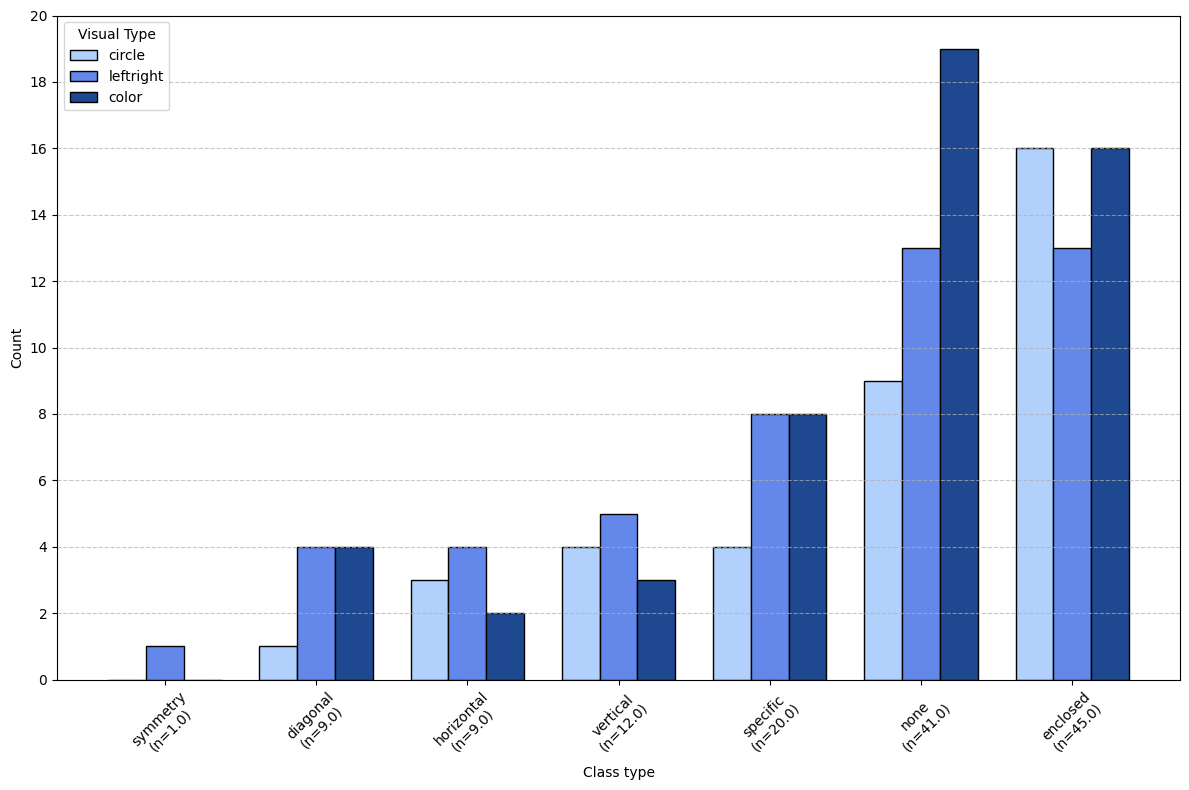}
        \caption{\gpt free response results by rule class.}
        \label{fig:free_class_gpt}
    \end{minipage} \hfill
    \begin{minipage}[b]{0.45\textwidth}
        \centering
        \includegraphics[width=\linewidth]{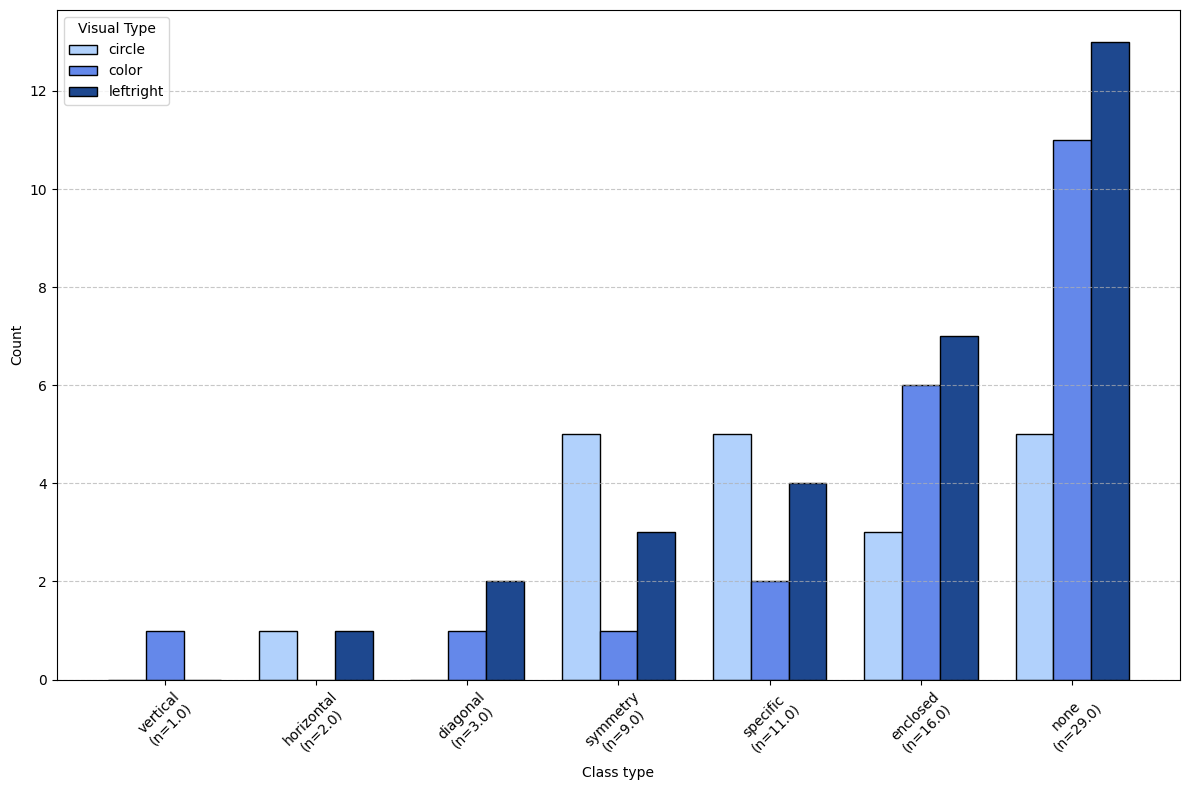}
        \caption{\gemini free response results by rule class.}
        \label{fig:free_class_gemini}
    \end{minipage}
\end{figure}

A passing pattern is true of all included characters and false of all excluded characters. In borderline cases, we ask whether the rule applies equally well to a character from each set; if so, it fails, since a human Glyph player would be unlikely to pick only the correct characters. We measure inter-annotator agreement using Cohen's $\kappa$. Figures \ref{fig:free_class_gpt} and \ref{fig:free_class_gemini} show the correct patterns generated by \gpt and \gemini in the free response task, grouped by the cross-linguistically common pattern categories from Section \ref{sec:cats}.

\subsection{Explanation Prompting}

\begin{figure*}[h]
    \fbox{%
        \begin{minipage}{0.95\linewidth}
            {\centering\includegraphics[width=0.5\linewidth]{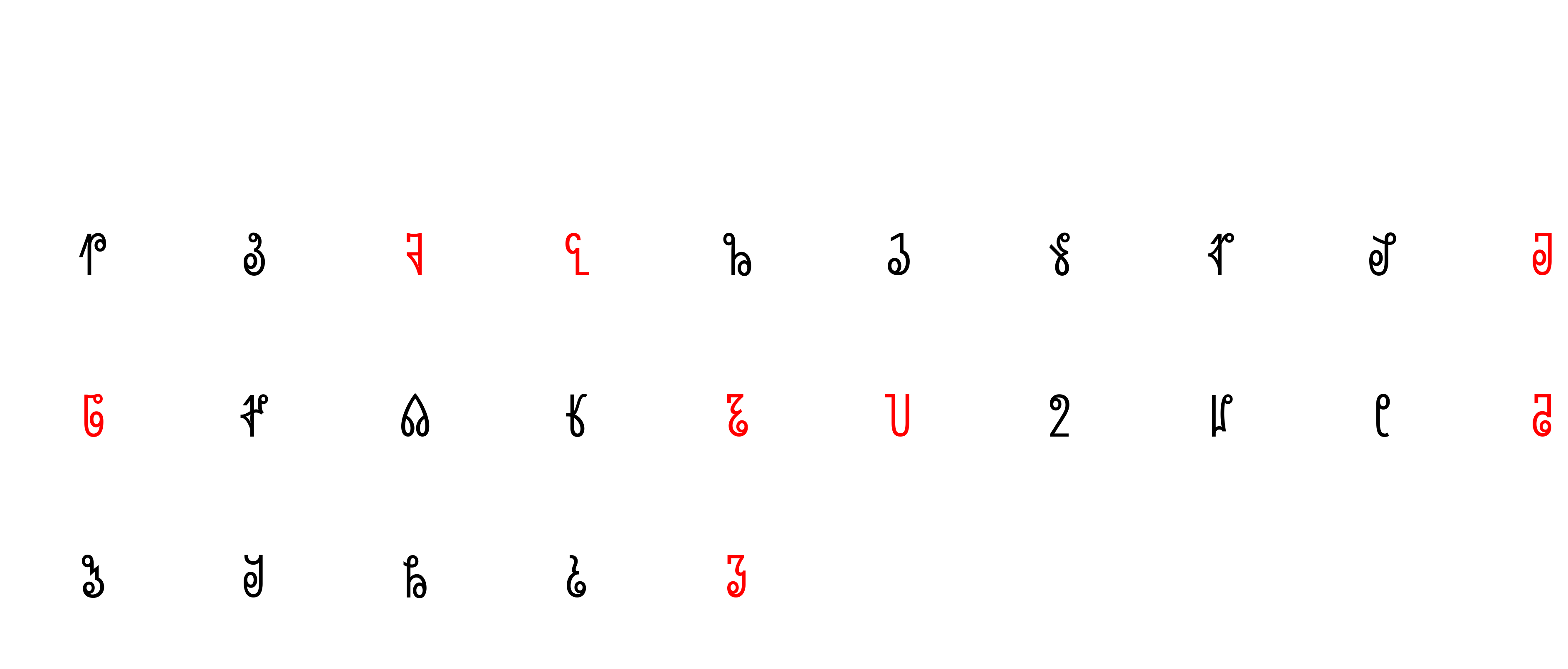}\\}
            \textbf{Groundtruth}:"B.Characters that contain at least one right angle."\\
            \textbf{\gemini Response}:"The answer is **D**.

            * **A. Characters that contain a circular element.**  Incorrect, while some red characters contain circles, some do not.
            * **B. Characters that contain at least one right angle.** Incorrect, several red characters lack right angles.
            * **C. Characters that exceed the average size of the character set in height or width.** Incorrect, the red characters vary in size and are not all larger than the average.
            * **D. Characters that have a right turn at the bottom of a vertical line, like an L or reversed J.** Correct, all the red characters feature a right turn at the bottom of a vertical line. 
            
            Therefore, the characters colored red are those that have a right turn at the bottom of a vertical line, like an L or reversed J. This pattern is consistently observed in all the red characters. "\\
            \textbf{\gpt Response}:"The correct option is B. Characters that contain at least one right angle.

            Here's the step-by-step reasoning:
            
            A. Characters that contain a circular element.
            - Some of the red characters do not contain circular elements. For example, the red character in the first row, fourth column from the left, does not have a circular element. Therefore, this option is incorrect.
            
            B. Characters that contain at least one right angle.
            - All the red characters contain at least one right angle. For example, the red character in the first row, fourth column from the left, has a right angle. This option is correct.
            
            C. Characters that exceed the average size of the character set in height or width.
            - Not all red characters exceed the average size of the character set. For example, the red character in the second row, first column from the left, does not exceed the average size. Therefore, this option is incorrect.
            
            D. Characters that have a right turn at the bottom of a vertical line, like an L or reversed J.
            - Not all red characters have a right turn at the bottom of a vertical line. For example, the red character in the second row, fourth column from the left, does not have a right turn at the bottom of a vertical line. Therefore, this option is incorrect.
            
            Thus, the correct answer is B."\\
            
        \end{minipage}
    }
    \caption{Multiple Choice Explanation Example 1}
    \label{fig:explainationex1}
\end{figure*}

\begin{figure*}[h]
    \fbox{%
        \begin{minipage}{\linewidth}
            {\centering\includegraphics[width=0.4\linewidth]{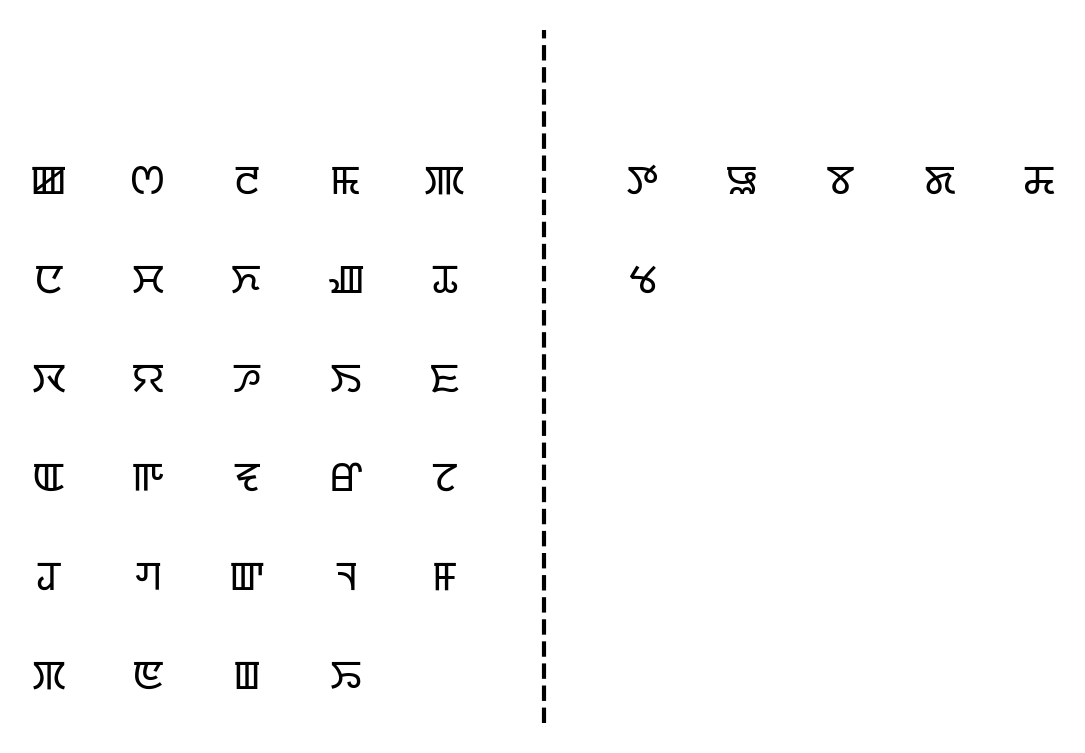}\\}
            \textbf{Groundtruth}:"D.Characters that have a closed loop or circular element, which may be crossed over itself in a swirl motion or formed by a line bisecting itself."\\
            \textbf{\gemini Response}:"The answer is **D. Characters that have a closed loop or circular element, which may be crossed over itself in a swirl motion or formed by a line bisecting itself.**

            Here's why:
            
            * **A. Characters that have parallel vertical lines or multiple vertical lines.**  While some characters on the right have vertical lines, this isn't a defining characteristic shared by all.
            * **B. Characters that have a large, open, vertical circle curve on the left side.** This describes some characters on the *left* side of the image, not the right.
            * **C. Characters that have a full-height straight line on the right side, ending in a vertical line perpendicular to a horizontal line at the base of the character.** This pattern doesn't accurately describe any of the characters on the right.
            
            **D is the most accurate:** Every character on the right side features a closed loop or circle:
            
            * Some loops are simple circles (like the third character).
            * Some are more complex, with lines crossing through (like the first and last characters).
            * The second character's loop is formed by a line bisecting itself. "\\
            \textbf{\gpt Response}:"The correct option is D.Let's analyze each option:

            A. Characters that have parallel vertical lines or multiple vertical lines.
            - The characters on the right side do not predominantly feature parallel vertical lines or multiple vertical lines. This option does not match.
            
            B. Characters that have a large, open, vertical circle curve on the left side.
            - The characters on the right side do not have a large, open, vertical circle curve on the left side. This option does not match.
            
            C. Characters that have a full-height straight line on the right side, ending in a vertical line perpendicular to a horizontal line at the base of the character.
            - The characters on the right side do not consistently have a full-height straight line on the right side ending in a vertical line perpendicular to a horizontal line at the base. This option does not match.
            
            D. Characters that have a closed loop or circular element, which may be crossed over itself in a swirl motion or formed by a line bisecting itself.
            - The characters on the right side do have closed loops or circular elements, some of which are crossed over themselves or formed by a line bisecting itself. This option matches."\\
            
        \end{minipage}
    }
    \caption{Multiple Choice Explanation Example 2}
    \label{fig:explainationex2}
\end{figure*}

\begin{figure*}[h]
    \fbox{%
        \begin{minipage}{\linewidth}
            {\centering\includegraphics[width=0.4\linewidth]{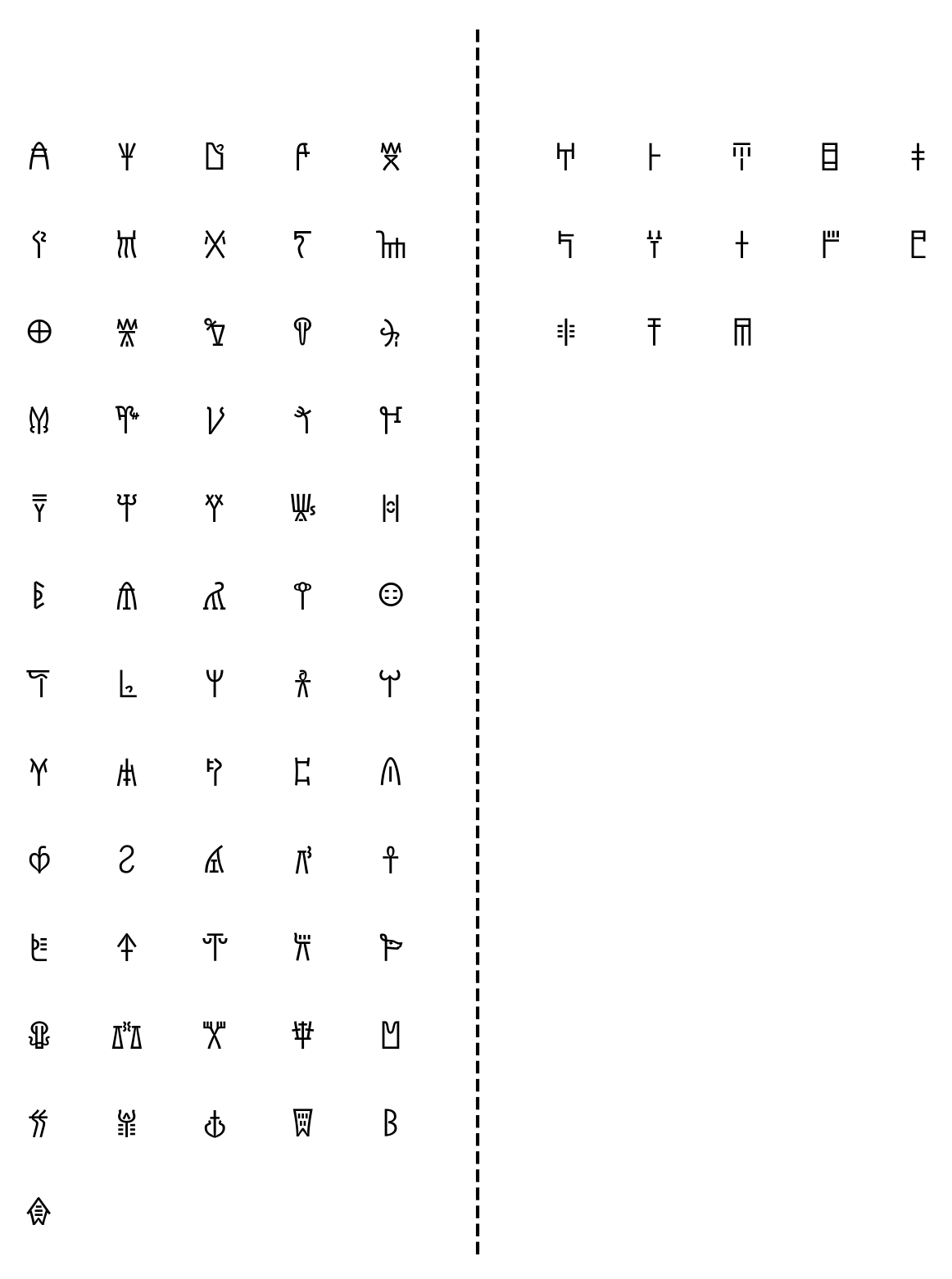}\\}
            \textbf{Groundtruth}:"A. Characters that are made up of only horizontal and vertical lines with no curves or diagonal lines and only right angles."\\
                        \textbf{\gemini Response}:"B. Characters that have at least one enclosed space or loop.

            Here's why:
            
            * **A. Characters that are made up of only horizontal and vertical lines with no curves or diagonal lines and only right angles.**  This is incorrect. Many characters on the right side have curves or diagonal lines (e.g.,"B").
            
            * **B. Characters that have at least one enclosed space or loop.** This is the correct answer. Every character on the right side has at least one enclosed space or loop.
            
            * **C. Characters that are similar to an F, with a vertical segment on the right side and two small oblique segments descending to the right.** This is incorrect.  While some characters on the right side might share similarities with an "F", this description doesn't fit all of them.
            
            * **D. Characters that do not have an overhead horizontal straight line.** This is incorrect. Some characters on the right side do have an overhead horizontal line (e.g., "B", "E")."\\
            \textbf{\gpt Response}:"The correct option is A.
            
            Let's analyze each option step-by-step:
            
            A. Characters that are made up of only horizontal and vertical lines with no curves or diagonal lines and only right angles.
            - The characters on the right side of the image are composed of only horizontal and vertical lines, with no curves or diagonal lines, and they all have right angles. This matches the description perfectly.
            
            B. Characters that have at least one enclosed space or loop.
            - Not all characters on the right side have enclosed spaces or loops. For example, the character in the second row, first column on the right side does not have any enclosed space or loop.
            
            C. Characters that are similar to an F, with a vertical segment on the right side and two small oblique segments descending to the right.
            - None of the characters on the right side resemble an F with a vertical segment on the right side and two small oblique segments descending to the right.
            
            D. Characters that do not have an overhead horizontal straight line.
            - Some characters on the right side do have an overhead horizontal straight line. For example, the character in the first row, first column on the right side has an overhead horizontal straight line.
            
            Therefore, the correct answer is A."\\

        \end{minipage}
    }
    \caption{Multiple Choice Explanation Example 3}
    \label{fig:explainationex3}
\end{figure*}

\begin{figure*}[h]
    \fbox{%
        \begin{minipage}{\linewidth}
            {\centering\includegraphics[width=0.4\linewidth]{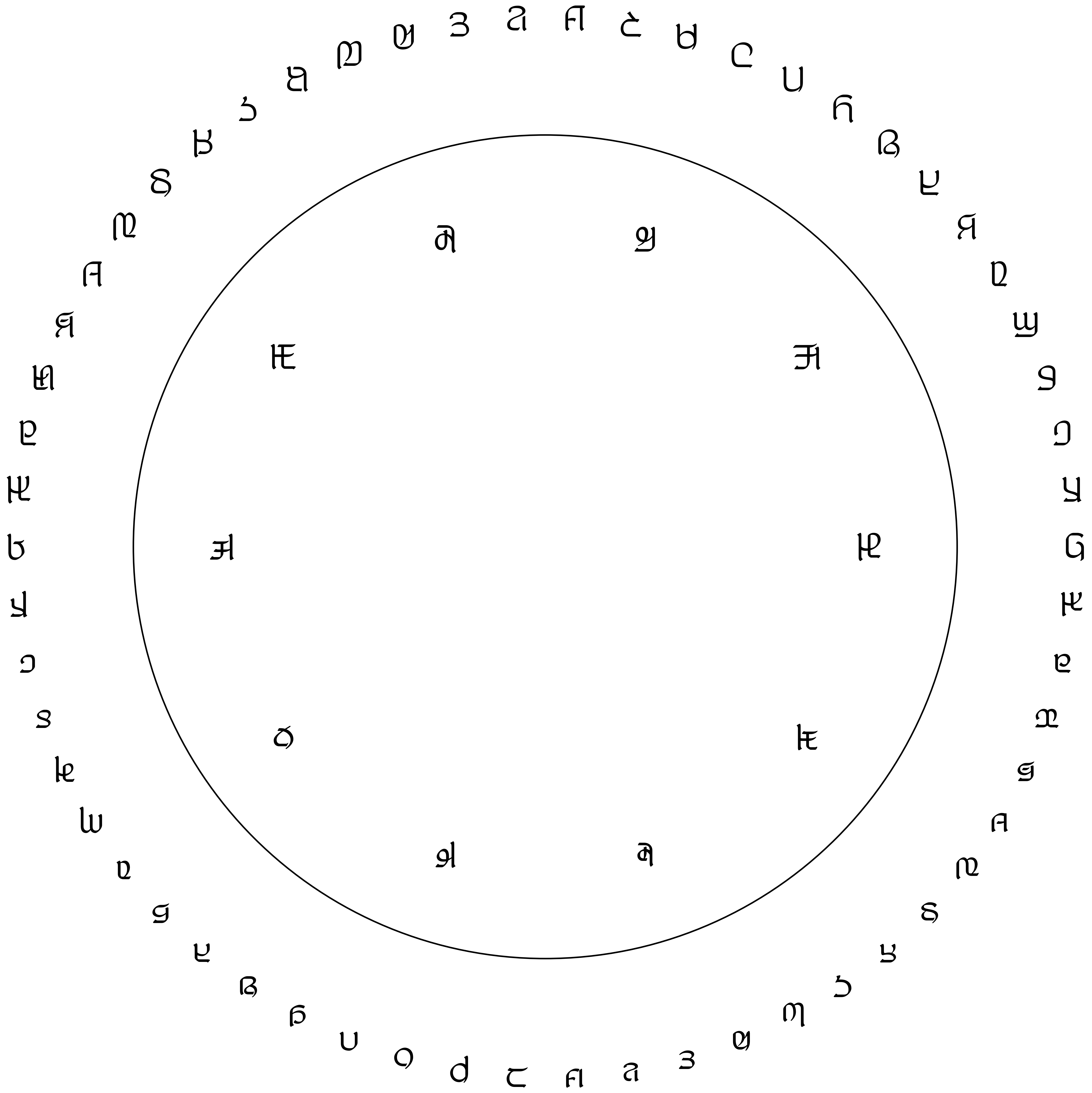}\\}
            \textbf{Groundtruth}:"C. Characters that have lines that cross each other."\\
            \textbf{\gemini Response}:"The answer is **D**.

            * **A. Characters that contain at least one straight vertical line.**  This is incorrect because some characters inside the circle, like "H", "D", and "O", have vertical lines, but some characters outside the circle, like "I", "T", and "L", also have vertical lines. 
            * **B. Characters that have a straight horizontal stroke at the bottom or a flat bottom, or are used in font encoding.** This is incorrect because characters like "H", "O", and "B" inside the circle do not have horizontal lines at the bottom and font encoding is not a relevant factor. 
            * **C. Characters that have lines that cross each other.** This is incorrect because "B", "D", "O", and "C" are inside the circle, while "F" is outside the circle, all of them having crossing lines. 
            * **D. Characters that have at least one completely enclosed space or closed loop.** This is correct because all characters inside the circle have closed loops or enclosed spaces. There are no characters outside the circle with enclosed spaces. 
            
            Therefore, the correct answer is **D**."\\
            \textbf{\gpt Response}:"The correct option is C. Characters that have lines that cross each other.

            Step-by-step reasoning through all options:
            
            A. Characters that contain at least one straight vertical line.
            - Many characters both inside and outside the circle contain straight vertical lines. For example, the character "E" inside the circle and the character "L" outside the circle both have straight vertical lines. This option does not distinguish between the characters inside and outside the circle.
            
            B. Characters that have a straight horizontal stroke at the bottom or a flat bottom, or are used in font encoding.
            - This option is too broad and does not specifically apply to the characters inside the circle. Characters both inside and outside the circle have straight horizontal strokes at the bottom or flat bottoms.
            
            C. Characters that have lines that cross each other.
            - This option correctly identifies the characters inside the circle. All characters inside the circle have lines that cross each other, such as "E", "F", and "H". Characters outside the circle do not have lines that cross each other.
            
            D. Characters that have at least one completely enclosed space or closed loop.
            - This option does not apply to the characters inside the circle. Characters both inside and outside the circle have enclosed spaces or closed loops, such as "O" outside the circle and "E" inside the circle.
            
            Therefore, the correct answer is C. Characters that have lines that cross each other."\\
            
        \end{minipage}
    }
    \caption{Multiple Choice Explanation Example 4}
    \label{fig:explainationex4}
\end{figure*}

\begin{figure*}[h]
    \fbox{%
        \begin{minipage}{\linewidth}
            {\centering\includegraphics[width=0.35\linewidth]{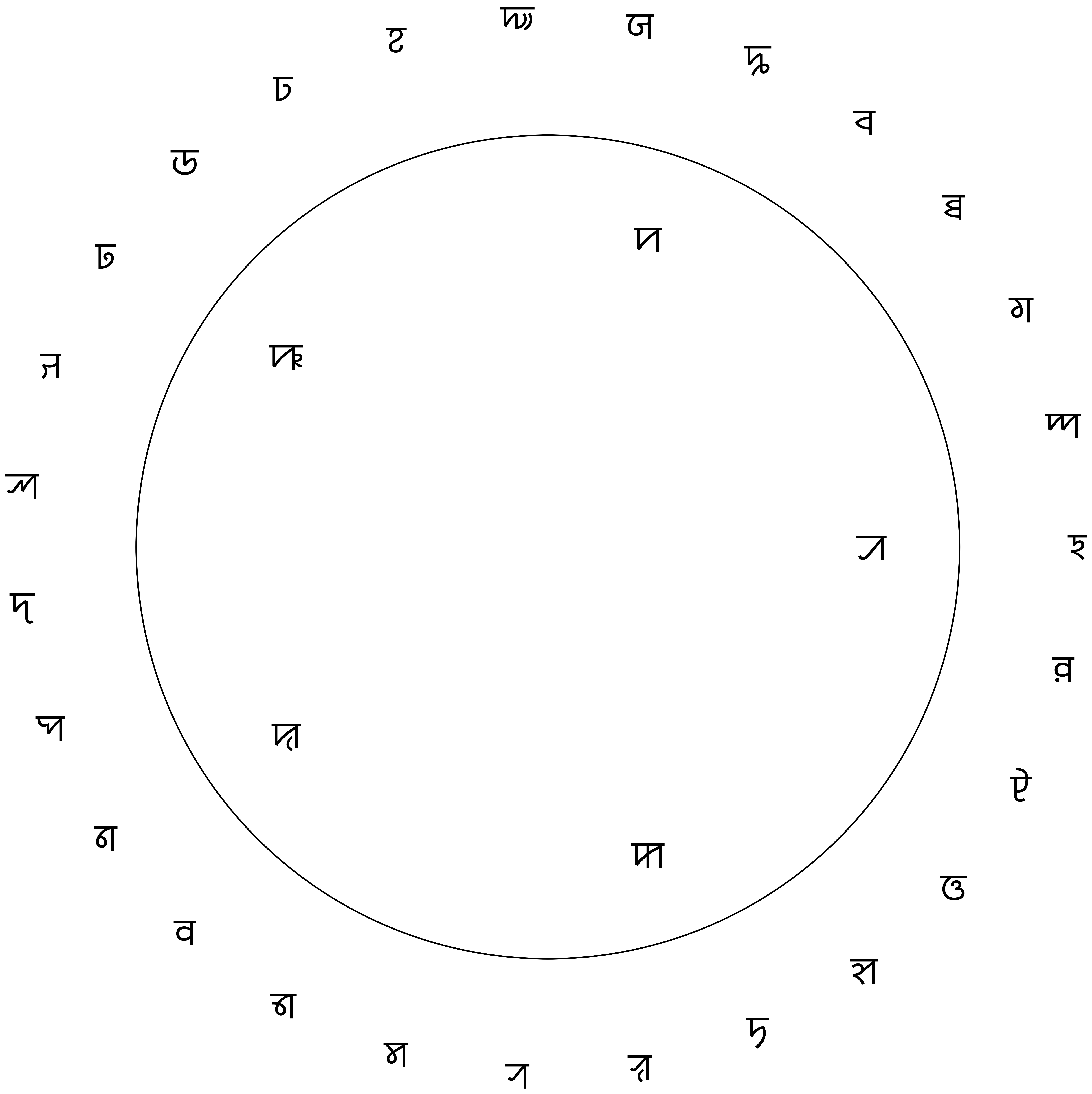}\\}
            \textbf{Groundtruth}:"D. Characters that contain a full-length diagonal line bottom left to top right."\\
            \textbf{\gemini response}:"The answer is **D**.

            Let's analyze each option:
            
            * **A. Characters that cannot be drawn in a single stroke or without going over part of an existing line.** This is not true for all characters inside the circle. For example, the character "{\dn m}" can be drawn in a single stroke.
            * **B. Characters that have horizontal symmetry.** This is not true for all characters inside the circle. For example, the character "{\dn m}" does not have horizontal symmetry.
            * **C. Characters that have a horizontal line at the bottom.** This is not true for all characters inside the circle. For example, the character "{\dn k}" does not have a horizontal line at the bottom.
            * **D. Characters that contain a full length diagonal line bottom left to top right.** This is true for all characters inside the circle. The characters  "{\dn k}", "{\dn m}", "{\dn g}", "{\dn y}", and "{\dn j}" all contain a full length diagonal line bottom left to top right.
            
            Therefore, the correct answer is **D**."\\
            \textbf{\gpt Response}:"The correct option is A. Characters that cannot be drawn in a single stroke or without going over part of an existing line.

            Step-by-step reasoning through all options:
            
            A. Characters that cannot be drawn in a single stroke or without going over part of an existing line.
            - Inside the circle, the characters are complex and require multiple strokes to draw without lifting the pen or going over an existing line. This matches the characters inside the circle.
            
            B. Characters that have horizontal symmetry.
            - Horizontal symmetry means the character looks the same when flipped horizontally. The characters inside the circle do not exhibit horizontal symmetry.
            
            C. Characters that have a horizontal line at the bottom.
            - The characters inside the circle do not consistently have a horizontal line at the bottom.
            
            D. Characters that contain a full-length diagonal line bottom left to top right.
            - The characters inside the circle do not consistently contain a full-length diagonal line from bottom left to top right.
            
            Therefore, the correct answer is A."\\
            
        \end{minipage}
    }
    \caption{Multiple Choice Explanation Example 5}
    \label{fig:explainationex5}
\end{figure*}

To elicit model explanations for our manual error analysis, we prompt \gemini and \gpt for an explanation after their answer: \textit{Reply with the letter of the correct option, and then explain your step-by-step reasoning through all options}. Figure \ref{fig:explainationex1}, \ref{fig:explainationex2}, \ref{fig:explainationex3}, \ref{fig:explainationex4}, and \ref{fig:explainationex5} are five examples showing how the models go awry in their explanations. The answer choices can be seen in the models' response.

\subsection{Logistic Regression Results}\label{app:mem}

We test the statistical reliability of various factors that impact model performance using mixed-effects logistic regression models. We treat correctness as a binary outcome and include item random effects. We include fixed effects for \style, pattern length, pattern category (using the cross-linguistically common patterns discussed in Section \ref{sec:cats}), and writing system. We fit models to each VLM's results using the lme4 package in R~\citep{lme4}. Tables \ref{tab:mem_gemini}-\ref{tab:mem_molmod2} show the results. We split the main effects and language effects into separate tables for length reasons, but all effects were fitted as part of the same model.

\section{Use of Generative AI}\label{use-of-ai-assistants}

Some code for this paper was written with AI assistants enabled.

\clearpage

\begin{table}[t]
\begin{tabular}{|l|lll|}\hline
Fixed effects&$\widehat{\beta}$&$z$&$p$\\\hline
(Intercept)                  & 0.12 ($\pm$ 0.15)& 0.83 & 0.41 \\
color                  & 0.10 ($\pm$ 0.03)& 3.11 & \textbf{0.002}\\
leftright              & 0.09 ($\pm$ 0.03)& 2.74 & \textbf{0.006}\\\hline
length                       & 0.003 ($\pm$ 0.001)& 3.24 & \textbf{0.001}\\\hline
cross                   & -0.31 ($\pm$ 0.15)& -2.08 & \textbf{0.04}\\
diagonal                & 0.27 ($\pm$ 0.15)& 1.50 & 0.14 \\
enclosed                & 0.18 ($\pm$ 0.06)& 2.97 & \textbf{0.003}\\
horizontal              & -0.08 ($\pm$ 0.11)& -0.76 & 0.45 \\
mirror                  & -0.05 ($\pm$ 0.27)& -0.19 & 0.85 \\
specific                & 0.13 ($\pm$ 0.06)& 2.07 & \textbf{0.04}\\
stroke                  & 0.02 ($\pm$ 0.08)& 0.29 & 0.77 \\
symmetry                & -0.17 ($\pm$ 0.08)& -2.16 & \textbf{0.03}\\
vertical                & 0.28 ($\pm$ 0.11)& 2.64 & \textbf{0.009}\\\hline
\end{tabular}
\caption{Full logistic regression table for \gemini, main effects}\label{tab:mem_gemini}
\end{table}

\begin{table}[t]
\begin{tabular}{|l|lll|}\hline
Fixed effects&$\widehat{\beta}$&$z$&$p$\\\hline
(Intercept) & 0.27 ($\pm$ 0.15) & 1.84 & 0.066 \\
color & 0.06 ($\pm$ 0.03) & 2.05 & \textbf{0.041}\\
leftright & 0.07 ($\pm$ 0.03) & 2.15 & \textbf{0.032}\\\hline
length & 0.003 ($\pm$ 0.001) & 3.12 & \textbf{0.002}\\\hline
cross & 0.13 ($\pm$ 0.15) & 0.91 & 0.37 \\
diagonal & 0.46 ($\pm$ 0.15) & 3.07 & \textbf{0.002}\\
enclosed & 0.24 ($\pm$ 0.06) & 4.17 & \textbf{$<$0.001}\\
horizontal & -0.23 ($\pm$ 0.11) & -2.20 & \textbf{0.028}\\
mirror & -0.24 ($\pm$ 0.27) & -0.89 & 0.38 \\
specific & 0.37 ($\pm$ 0.06) & 5.78 & \textbf{$<$0.001}\\
stroke & -0.13 ($\pm$ 0.08) & -1.56 & 0.12 \\
symmetry & -0.09 ($\pm$ 0.08) & -1.16 & 0.25 \\
vertical & 0.19 ($\pm$ 0.11) & 1.75 & 0.08 \\\hline
\end{tabular}
\caption{Full logistic regression table for \gpt, main effects}\label{tab:mem_gpt}
\end{table}

\clearpage

\begin{table}[t]
\begin{tabular}{|l|lll|}\hline
Fixed effects&$\widehat{\beta}$&$z$&$p$\\\hline
Avestan                & 0.01 ($\pm$ 0.19)& 0.08 & 0.94 \\
Bamum                  & -0.01 ($\pm$ 0.20)& -0.05 & 0.96 \\
Bassa Vah              & 0.17 ($\pm$ 0.18)& 0.91 & 0.37 \\
Batak                  & 0.04 ($\pm$ 0.19)& 0.19 & 0.85 \\
Buginese               & 0.36 ($\pm$ 0.24)& 1.50 & 0.14 \\
Cherokee               & 0.24 ($\pm$ 0.18)& 1.32 & 0.19 \\
Gothic                 & -0.03 ($\pm$ 0.22)& -0.16 & 0.88 \\
Grantha                & -0.19 ($\pm$ 0.17)& -1.16 & 0.25 \\
Hangul                 & 0.15 ($\pm$ 0.18)& 0.81 & 0.42 \\
Kayah Li               & 0.11 ($\pm$ 0.18)& 0.59 & 0.55 \\
Khudawadi              & -0.05 ($\pm$ 0.22)& -0.23 & 0.82 \\
Lao                    & 0.14 ($\pm$ 0.17)& 0.82 & 0.42 \\
Lepcha                 & 0.32 ($\pm$ 0.18)& 1.77 & 0.08 \\
Linear B               & 0.01 ($\pm$ 0.18)& 0.05 & 0.96 \\
Mandaic                & 0.07 ($\pm$ 0.18)& 0.39 & 0.69 \\
Meetei Mayek           & 0.17 ($\pm$ 0.17)& 1.04 & 0.30 \\
Meroitic Cursive       & 0.05 ($\pm$ 0.24)& 0.19 & 0.85 \\
Mro                    & 0.16 ($\pm$ 0.20)& 0.80 & 0.42 \\
Multani                & 0.10 ($\pm$ 0.17)& 0.59 & 0.56 \\
Nabataean              & 0.10 ($\pm$ 0.19)& 0.55 & 0.58 \\
New Tai Lue            & 0.38 ($\pm$ 0.19)& 1.94 & 0.05 \\
Nyiakeng Puachue Hmong& 0.05 ($\pm$ 0.19)& 0.27 & 0.79 \\
Ogham                  & -0.05 ($\pm$ 0.20)& -0.25 & 0.80 \\
Ol Chiki               & 0.22 ($\pm$ 0.19)& 1.17 & 0.24 \\
Old Permic             & 0.15 ($\pm$ 0.17)& 0.85 & 0.40 \\
Osmanya                & 0.07 ($\pm$ 0.27)& 0.25 & 0.80 \\
Pau Cin Hau            & 0.23 ($\pm$ 0.17)& 1.38 & 0.17 \\
Phoenician             & 0.24 ($\pm$ 0.20)& 1.16 & 0.25 \\
Pracalit               & 0.02 ($\pm$ 0.18)& 0.10 & 0.92 \\
Psalter Pahlavi        & 0.38 ($\pm$ 0.20)& 1.88 & 0.06 \\
Runic                  & -0.02 ($\pm$ 0.17)& -0.11 & 0.92 \\
Shavian                & 0.007 ($\pm$ 0.18)& 0.04 & 0.97 \\
Sora Sompeng           & 0.07 ($\pm$ 0.17)& 0.40 & 0.69 \\
Soyombo                & -0.04 ($\pm$ 0.17)& -0.26 & 0.80 \\
Syloti Nagri           & -0.03 ($\pm$ 0.18)& -0.14 & 0.89 \\
Tagbanwa               & 0.58 ($\pm$ 0.20)& 2.86 & \textbf{0.005}\\
Tai Viet               & 0.11 ($\pm$ 0.19)& 0.58 & 0.56 \\
Warang Citi            & 0.26 ($\pm$ 0.20)& 1.34 & 0.18 \\
Zanabazar Square       & 0.19 ($\pm$ 0.17)& 1.14 & 0.26 \\\hline
\end{tabular}
\caption{Full logistic regression table for \gemini, language effects}\label{tab:mem2_gemini}
\end{table}

\clearpage

\begin{table}[t]
\begin{tabular}{|l|lll|}\hline
Fixed effects&$\widehat{\beta}$&$z$&$p$\\\hline
Avestan & -0.08 ($\pm$ 0.19) & -0.43 & 0.67 \\
Bamum & 0.11 ($\pm$ 0.20) & 0.54 & 0.59 \\
Bassa Vah & 0.23 ($\pm$ 0.18) & 1.26 & 0.21 \\
Batak & 0.25 ($\pm$ 0.19) & 1.34 & 0.18 \\
Buginese & 0.12 ($\pm$ 0.24) & 0.51 & 0.61 \\
Cherokee & 0.21 ($\pm$ 0.18) & 1.17 & 0.24 \\
Gothic & 0.23 ($\pm$ 0.22) & 1.04 & 0.30 \\
Grantha & -0.40 ($\pm$ 0.17) & -2.40 & \textbf{0.017}\\
Hangul & 0.20 ($\pm$ 0.18) & 1.12 & 0.27 \\
Kayah Li & -0.05 ($\pm$ 0.18) & -0.28 & 0.78 \\
Khudawadi & 0.17 ($\pm$ 0.22) & 0.77 & 0.44 \\
Lao & -0.06 ($\pm$ 0.17) & -0.37 & 0.71 \\
Lepcha & 0.35 ($\pm$ 0.18) & 1.91 & 0.057 \\
Linear B & 0.09 ($\pm$ 0.18) & 0.52 & 0.60 \\
Mandaic & -0.04 ($\pm$ 0.18) & -0.20 & 0.84 \\
Meetei Mayek & 0.02 ($\pm$ 0.17) & 0.11 & 0.91 \\
Meroitic Cursive & -0.52 ($\pm$ 0.24) & -2.18 & \textbf{0.03}\\
Mro & 0.22 ($\pm$ 0.19) & 1.16 & 0.25 \\
Multani & 0.11 ($\pm$ 0.17) & 0.65 & 0.52 \\
Nabataean & -0.08 ($\pm$ 0.19) & -0.41 & 0.68 \\
New Tai Lue & 0.30 ($\pm$ 0.19) & 1.55 & 0.12 \\
Nyiakeng Puachue Hmong& -0.05 ($\pm$ 0.19) & -0.24 & 0.81 \\
Ogham & 0.10 ($\pm$ 0.20) & 0.49 & 0.63 \\
Ol Chiki & -0.05 ($\pm$ 0.19) & -0.27 & 0.79 \\
Old Permic & -0.03 ($\pm$ 0.17) & -0.18 & 0.86 \\
Osmanya & 0.28 ($\pm$ 0.27) & 1.02 & 0.31 \\
Pau Cin Hau & 0.06 ($\pm$ 0.17) & 0.38 & 0.71 \\
Phoenician & 0.07 ($\pm$ 0.20) & 0.36 & 0.72 \\
Pracalit & -0.20 ($\pm$ 0.18) & -1.12 & 0.27 \\
Psalter Pahlavi & 0.20 ($\pm$ 0.20) & 0.98 & 0.33 \\
Runic & 0.08 ($\pm$ 0.17) & 0.48 & 0.63 \\
Shavian & 0.07 ($\pm$ 0.18) & 0.40 & 0.69 \\
Sora Sompeng & -0.09 ($\pm$ 0.17) & -0.52 & 0.61 \\
Soyombo & -0.24 ($\pm$ 0.17) & -1.43 & 0.15 \\
Syloti Nagri & -0.19 ($\pm$ 0.18) & -1.05 & 0.30 \\
Tagbanwa & 0.21 ($\pm$ 0.20) & 1.03 & 0.31 \\
Tai Viet & 0.07 ($\pm$ 0.19) & 0.39 & 0.70 \\
Warang Citi & 0.19 ($\pm$ 0.19) & 0.96 & 0.34 \\
Zanabazar Square & 0.03 ($\pm$ 0.17) & 0.19 & 0.85 \\
\hline
\end{tabular}
\caption{Full logistic regression table for \gpt, language effects}\label{tab:mem_gpt2}
\end{table}

\clearpage

\begin{table}[t]
\begin{tabular}{|l|lll|}\hline
Fixed effects&$\widehat{\beta}$&$z$&$p$\\\hline
(Intercept)                   & 0.25 ($\pm$ 0.13)  & 1.96  & 0.05\\
color                   & -0.038 ($\pm$ 0.031) & -1.20 & 0.23 \\
leftright               & -0.057 ($\pm$ 0.031) & -1.80 & 0.072\\\hline
length                        & 0.0019 ($\pm$ 0.00068) & 2.85 & \textbf{0.005}\\\hline
cross                    & -0.11 ($\pm$ 0.13)   & -0.84 & 0.40 \\
diagonal                 & 0.12 ($\pm$ 0.13)    & 0.92  & 0.36 \\
enclosed                 & 0.051 ($\pm$ 0.050)  & 1.03  & 0.30 \\
horizontal               & -0.097 ($\pm$ 0.089) & -1.09 & 0.28 \\
mirror                   & 0.23 ($\pm$ 0.23)    & 1.00  & 0.32 \\
specific                 & 0.48 ($\pm$ 0.053)   & 9.02  & \textbf{$<$ 0.001}\\
stroke                   & -0.097 ($\pm$ 0.068) & -1.43 & 0.15 \\
symmetry                 & 0.13 ($\pm$ 0.068)   & 1.97  & 0.05\\
vertical                 & 0.28 ($\pm$ 0.090)   & 3.15  & \textbf{0.002}\\\hline
\end{tabular}
\caption{Full logistic regression table for \idefics, main effects}\label{tab:mem_idefics}
\end{table}

\begin{table}[t]
\begin{tabular}{|l|lll|}\hline
Fixed effects&$\widehat{\beta}$&$z$&$p$\\\hline
(Intercept)                  & 0.21 ($\pm$ 0.13) & 1.67 & 0.10\\
color                  & 0.08 ($\pm$ 0.03) & 2.55 & \textbf{0.01} \\
leftright              & 0.04 ($\pm$ 0.03) & 1.13 & 0.26\\\hline
length                       & 0.002 ($\pm$ \textless 0.001) & 3.09 & \textbf{0.002}\\\hline
cross                   & -0.11 ($\pm$ 0.13) & -0.86 & 0.39\\
diagonal                & 0.06 ($\pm$ 0.13) & 0.46 & 0.64\\
enclosed                & 0.04 ($\pm$ 0.05) & 0.75 & 0.45\\
horizontal              & -0.10 ($\pm$ 0.09) & -1.07 & 0.28\\
mirror                  & -0.03 ($\pm$ 0.23) & -0.14 & 0.89\\
specific                & 0.34 ($\pm$ 0.05) & 6.24 & \textbf{\textless 0.001}\\
stroke                  & -0.06 ($\pm$ 0.07) & -0.87 & 0.39\\
symmetry                & -0.15 ($\pm$ 0.07) & -2.10 & \textbf{0.04}\\
vertical                & 0.12 ($\pm$ 0.09) & 1.27& 0.20\\\hline
\end{tabular}
\caption{Full logistic regression table for \ideficsN, main effects}\label{tab:mem_idefics3}
\end{table}

\clearpage

\begin{table}[t]
\begin{tabular}{|l|lll|}\hline
Fixed effects&$\widehat{\beta}$&$z$&$p$\\\hline
Avestan                 & -0.13 ($\pm$ 0.16)   & -0.85 & 0.40 \\
Bamum                   & -0.12 ($\pm$ 0.17)   & -0.73 & 0.47 \\
Bassa Vah               & -0.094 ($\pm$ 0.15)  & -0.61 & 0.54 \\
Batak                   & -0.31 ($\pm$ 0.16)   & -1.95 & 0.052\\
Buginese                & 0.027 ($\pm$ 0.20)   & 0.14  & 0.89 \\
Cherokee                & -0.13 ($\pm$ 0.15)   & -0.87 & 0.39 \\
Gothic                  & -0.0029 ($\pm$ 0.18) & -0.016 & 0.99 \\
Grantha                 & -0.23 ($\pm$ 0.14)   & -1.64 & 0.10 \\
Hangul                  & -0.0026 ($\pm$ 0.16) & -0.017 & 0.99 \\
Kayah Li                & -0.15 ($\pm$ 0.15)   & -0.98 & 0.33 \\
Khudawadi               & -0.020 ($\pm$ 0.18)  & -0.11 & 0.91 \\
Lao                     & 0.11 ($\pm$ 0.14)    & 0.75  & 0.45 \\
Lepcha                  & -0.17 ($\pm$ 0.15)   & -1.08 & 0.28 \\
Linear B                & -0.18 ($\pm$ 0.15)   & -1.19 & 0.24 \\
Mandaic                 & -0.088 ($\pm$ 0.15)  & -0.57 & 0.57 \\
Meetei Mayek            & 0.071 ($\pm$ 0.14)   & 0.50  & 0.62 \\
Meroitic Cursive        & 0.052 ($\pm$ 0.20)   & 0.26  & 0.80 \\
Mro                     & -0.10 ($\pm$ 0.16)   & -0.62 & 0.54 \\
Multani                 & -0.26 ($\pm$ 0.15)   & -1.78 & 0.08\\
Nabataean               & -0.069 ($\pm$ 0.16)  & -0.44 & 0.66 \\
New Tai Lue             & 0.14 ($\pm$ 0.16)    & 0.86  & 0.39 \\
Nyiakeng Puachue Hmong& -0.11 ($\pm$ 0.16)   & -0.69 & 0.49 \\
Ogham                   & -0.16 ($\pm$ 0.17)   & -0.95 & 0.35 \\
Ol Chiki                & -0.060 ($\pm$ 0.16)  & -0.38 & 0.71 \\
Old Permic              & -0.13 ($\pm$ 0.14)   & -0.87 & 0.38 \\
Osmanya                 & -0.10 ($\pm$ 0.23)   & -0.45 & 0.65 \\
Pau Cin Hau             & -0.17 ($\pm$ 0.14)   & -1.23 & 0.22 \\
Phoenician              & -0.33 ($\pm$ 0.17)   & -1.94 & 0.054\\
Pracalit                & -0.11 ($\pm$ 0.15)   & -0.73 & 0.47 \\
Psalter Pahlavi         & -0.13 ($\pm$ 0.17)   & -0.76 & 0.45 \\
Runic                   & -0.17 ($\pm$ 0.14)   & -1.23 & 0.22 \\
Shavian                 & 0.076 ($\pm$ 0.15)   & 0.50  & 0.62 \\
Sora Sompeng            & -0.10 ($\pm$ 0.14)   & -0.69 & 0.49 \\
Soyombo                 & -0.18 ($\pm$ 0.14)   & -1.26 & 0.21 \\
Syloti Nagri            & -0.014 ($\pm$ 0.15)  & -0.093 & 0.93 \\
Tagbanwa                & -0.097 ($\pm$ 0.17)  & -0.57 & 0.57 \\
Tai Viet                & -0.037 ($\pm$ 0.16)  & -0.24 & 0.81 \\
Warang Citi             & 0.046 ($\pm$ 0.16)   & 0.28  & 0.78 \\
Zanabazar Square        & 0.039 ($\pm$ 0.14)   & 0.28  & 0.78 \\
\hline
\end{tabular}
\caption{Full logistic regression table for \idefics, language effects}\label{tab:mem_idefics2}
\end{table}

\clearpage

\begin{table}[t]
\begin{tabular}{|l|lll|}\hline
Fixed effects&$\widehat{\beta}$&$z$&$p$\\\hline
Avestan                & -0.11 ($\pm$ 0.16) & -0.66 & 0.51\\
Bamum                  & -0.24 ($\pm$ 0.17) & -1.35& 0.18\\
Bassa Vah              & -0.05 ($\pm$ 0.16) & -0.31 & 0.76\\
Batak                  & -0.12 ($\pm$ 0.16) & -0.75 & 0.46\\
Buginese               & 0.31 ($\pm$ 0.20) & 1.54 & 0.13\\
Cherokee               & 0.22 ($\pm$ 0.15) & 1.45 & 0.15\\
Gothic                 & -0.07 ($\pm$ 0.19) & -0.37 & 0.71\\
Grantha                & -0.19 ($\pm$ 0.14) & -1.37 & 0.17\\
Hangul                 & -0.04 ($\pm$ 0.16) & -0.23 & 0.82\\
Kayah Li               & -0.02 ($\pm$ 0.15) & -0.15 & 0.88\\
Khudawadi              & -0.03 ($\pm$ 0.19) & -0.16 & 0.88\\
Lao                    & 0.05 ($\pm$ 0.15) & 0.36 & 0.72\\
Lepcha                 & -0.10 ($\pm$ 0.16) & -0.66 & 0.51\\
Linear B               & -0.22 ($\pm$ 0.15) & -1.48 & 0.14\\
Mandaic                & -0.12 ($\pm$ 0.16) & -0.78 & 0.44\\
Meetei Mayek           & -0.03 ($\pm$ 0.14) & -0.20 & 0.84\\
Meroitic Cursive       & -0.11 ($\pm$ 0.21) & -0.52 & 0.60\\
Mro                    & -0.23 ($\pm$ 0.17) & -1.35 & 0.18 \\
Multani                & -0.07 ($\pm$ 0.15) & -0.44 & 0.66 \\
Nabataean              & -0.11 ($\pm$ 0.16) & -0.67 & 0.51\\
New Tai Lue            & 0.15 ($\pm$ 0.17) & 0.87 & 0.38\\
Nyiakeng Puachue Hmong& 0.1016 ($\pm$ 0.16) & 0.63 & 0.53\\
Ogham                  & 0.05 ($\pm$ 0.17) & 0.27 & 0.79\\
Ol Chiki               & -0.12 ($\pm$ 0.16) & -0.77 & 0.45\\
Old Permic             & -0.14 ($\pm$ 0.15) & -0.93 & 0.35\\
Osmanya                & -0.32 ($\pm$ 0.23) & -1.35 & 0.18 \\
Pau Cin Hau            & 0.02 ($\pm$ 0.14) & 0.14 & 0.89 \\
Phoenician             & -0.26 ($\pm$ 0.17) & -1.47 & 0.14\\
Pracalit               & -0.01 ($\pm$ 0.15) & -0.08 & 0.93\\
Psalter Pahlavi        & -0.14 ($\pm$ 0.17) & -0.83 & 0.41\\
Runic                  & 0.05 ($\pm$ 0.14) & 0.38 & 0.70 \\
Shavian                & -0.10 ($\pm$ 0.16) & -0.67 & 0.50 \\
Sora Sompeng           & -0.10 ($\pm$ 0.15) & -0.67 & 0.51\\
Soyombo                & -0.28 ($\pm$ 0.14) & -1.97 & 0.05\\
Syloti Nagri           & -0.17 ($\pm$ 0.15) & -1.11 & 0.27 \\
Tagbanwa               & -0.15 ($\pm$ 0.17) & -0.86 & 0.39 \\
Tai Viet               & 0.11 ($\pm$ 0.16) & 0.66 & 0.51 \\
Warang Citi            & 0.25 ($\pm$ 0.17) & 1.51 & 0.13\\
Zanabazar Square       & -0.04 ($\pm$ 0.14) & -0.31 & 0.76\\
\hline
\end{tabular}
\caption{Full logistic regression table for \ideficsN, language effects}\label{tab:mem_idefics32}
\end{table}

\clearpage

\begin{table}[t]
\begin{tabular}{|l|lll|}\hline
Fixed effects&$\widehat{\beta}$&$z$&$p$\\\hline
(Intercept)                   & 0.33 ($\pm$ 0.14)  & 2.5  & \textbf{0.02}\\
color                   & 0.050 ($\pm$ 0.032)  & 1.6  & 0.11  \\
leftright               & 0.022 ($\pm$ 0.032)  & 0.70  & 0.49  \\\hline
length                        & 0.0014 ($\pm$ \textless 0.001)  & 1.9  & 0.05\\\hline
cross                    & -0.18 ($\pm$ 0.14) & -1.3 & 0.18  \\
diagonal                 & 0.055 ($\pm$ 0.14)  & 0.40  & 0.69  \\
enclosed                 & 0.26 ($\pm$ 0.053)  & 4.8  & \textbf{\textless 0.001}\\
horizontal               & -0.031 ($\pm$ 0.095) & -0.32 & 0.75  \\
mirror                   & 0.52 ($\pm$ 0.25)  & 2.1  & \textbf{0.04}\\
specific                 & 0.37 ($\pm$ 0.057)  & 6.5  & \textbf{\textless 0.001}\\
stroke                   & 0.026 ($\pm$ 0.073) & 0.36 & 0.72  \\
symmetry                 & 0.0022 ($\pm$ 0.073) & 0.030 & 0.98  \\
vertical                 & 0.043 ($\pm$ 0.097) & 0.44 & 0.66  \\\hline
\end{tabular}
\caption{Full logistic regression table for \llava, main effects.}\label{tab:mem_llava}
\end{table}

\begin{table}[t]
\begin{tabular}{|l|lll|}\hline
Fixed effects&$\widehat{\beta}$&$z$&$p$\\\hline
(Intercept) & 0.17 ($\pm$ 0.12) & 1.48 & 0.14 \\
color & 0.01 ($\pm$ 0.03) & 0.30 & 0.77 \\
leftright & 0.03 ($\pm$ 0.03) & 1.08 & 0.28 \\\hline
length & 0.002 ($\pm$\textless 0.001) & 3.88 & \textbf{0.0001}\\\hline
cross & -0.24 ($\pm$ 0.12) & -2.04 & \textbf{0.04} \\
diagonal & -0.06 ($\pm$ 0.12) & -0.54 & 0.59 \\
enclosed & 0.02 ($\pm$ 0.05) & 0.39 & 0.70 \\
horizontal & -0.09 ($\pm$ 0.08) & -1.15 & 0.25 \\
mirror & -0.06 ($\pm$ 0.22) & -0.26 & 0.80 \\
specific & 0.17 ($\pm$ 0.05) & 3.34 & \textbf{0.001}\\
stroke & -0.15 ($\pm$ 0.06) & -2.37 & \textbf{0.02} \\
symmetry & -0.04 ($\pm$ 0.06) & -0.64 & 0.52 \\
vertical & 0.06 ($\pm$ 0.08) & 0.71 & 0.48 \\\hline
\end{tabular}
\caption{Full logistic regression table for \blip, main effects.}\label{tab:mem_blip}
\end{table}

\clearpage

\begin{table}[t]
\begin{tabular}{|l|lll|}\hline
Fixed effects&$\widehat{\beta}$&$z$&$p$\\\hline
Avestan                 & -0.29 ($\pm$ 0.17)  & -1.7  & 0.09\\
Bamum                   & -0.098 ($\pm$ 0.18) & -0.54 & 0.59\\
Bassa Vah               & -0.32 ($\pm$ 0.17)  & -1.9  & 0.05\\
Batak                   & -0.19 ($\pm$ 0.17)  & -1.1  & 0.25  \\
Buginese                & 0.023 ($\pm$ 0.21)  & 0.11  & 0.91  \\
Cherokee                & -0.16 ($\pm$ 0.16)  & -0.96 & 0.34  \\
Gothic                  & -0.39 ($\pm$ 0.20)  & -2.0  & 0.05\\
Grantha                 & -0.28 ($\pm$ 0.15)  & -1.9  & 0.06\\
Hangul                  & -0.27 ($\pm$ 0.17)  & -1.6  & 0.11  \\
Kayah Li                & -0.24 ($\pm$ 0.16)  & -1.5  & 0.14  \\
Khudawadi               & -0.24 ($\pm$ 0.20)  & -1.2  & 0.21  \\
Lao                     & -0.066 ($\pm$ 0.15) & -0.43 & 0.67  \\
Lepcha                  & -0.067 ($\pm$ 0.16) & -0.41 & 0.68  \\
Linear B                & -0.29 ($\pm$ 0.16)  & -1.8  & 0.07\\
Mandaic                 & -0.36 ($\pm$ 0.17)  & -2.2  & \textbf{0.03}\\
Meetei Mayek            & -0.068 ($\pm$ 0.15) & -0.45 & 0.65  \\
Meroitic Cursive        & -0.25 ($\pm$ 0.22)  & -1.1  & 0.25  \\
Mro                     & -0.32 ($\pm$ 0.18)  & -1.8  & 0.07\\
Multani                 & -0.27 ($\pm$ 0.16)  & -1.7  & 0.09\\
Nabataean               & -0.15 ($\pm$ 0.17)  & -0.86 & 0.39  \\
New Tai Lue             & -0.053 ($\pm$ 0.18) & -0.30 & 0.76  \\
Nyiakeng Puachue Hmong& -0.049 ($\pm$ 0.17) & -0.29 & 0.77  \\
Ogham                   & -0.075 ($\pm$ 0.18) & -0.41 & 0.68  \\
Ol Chiki                & -0.13 ($\pm$ 0.17)  & -0.76 & 0.45  \\
Old Permic              & -0.068 ($\pm$ 0.15) & -0.44 & 0.66  \\
Osmanya                 & -0.23 ($\pm$ 0.25)  & -0.92 & 0.36  \\
Pau Cin Hau             & -0.19 ($\pm$ 0.15)  & -1.2  & 0.22  \\
Phoenician              & -0.44 ($\pm$ 0.18)  & -2.4  & \textbf{0.02}\\
Pracalit                & -0.17 ($\pm$ 0.16)  & -1.1  & 0.29  \\
Psalter Pahlavi         & -0.069 ($\pm$ 0.18) & -0.37 & 0.71  \\
Runic                   & -0.25 ($\pm$ 0.15)  & -1.7  & 0.10  \\
Shavian                 & -0.13 ($\pm$ 0.16)  & -0.79 & 0.43  \\
Sora Sompeng            & -0.35 ($\pm$ 0.16)  & -2.3  & \textbf{0.03}\\
Soyombo                 & -0.49 ($\pm$ 0.15)  & -3.2  & \textbf{0.001}\\
Syloti Nagri            & -0.31 ($\pm$ 0.16)  & -1.9  & 0.05 \\
Tagbanwa                & -0.14 ($\pm$ 0.18)  & -0.79 & 0.43  \\
Tai Viet                & -0.13 ($\pm$ 0.17)  & -0.76 & 0.45  \\
Warang Citi             & -0.070 ($\pm$ 0.18) & -0.40 & 0.69  \\
Zanabazar Square        & -0.0030 ($\pm$ 0.15) & -0.020 & 0.98  \\
\hline
\end{tabular}
\caption{Full logistic regression table for \llava, language effects}\label{tab:mem_llava2}
\end{table}

\clearpage

\begin{table}[t]
\begin{tabular}{|l|lll|}\hline
Fixed effects&$\widehat{\beta}$&$z$&$p$\\\hline
Avestan & 0.04 ($\pm$ 0.15) & 0.27 & 0.78 \\
Bamum & 0.05 ($\pm$ 0.16) & 0.29 & 0.77 \\
Bassa Vah & -0.14 ($\pm$ 0.14) & -1.01 & 0.32 \\
Batak & -0.07 ($\pm$ 0.15) & -0.49 & 0.63 \\
Buginese & 0.36 ($\pm$ 0.19) & 1.95 & 0.05 \\
Cherokee & -0.24 ($\pm$ 0.14) & -1.68 & 0.09 \\
Gothic & -0.10 ($\pm$ 0.17) & -0.58 & 0.56 \\
Grantha & -0.16 ($\pm$ 0.13) & -1.23 & 0.22 \\
Hangul & -0.15 ($\pm$ 0.15) & -1.01 & 0.31 \\
Kayah Li & -0.15 ($\pm$ 0.14) & -1.11 & 0.27 \\
Khudawadi & -0.01 ($\pm$ 0.17) & -0.08 & 0.94 \\
Lao & -0.22 ($\pm$ 0.13) & -1.61 & 0.11 \\
Lepcha & -0.02 ($\pm$ 0.14) & -0.13 & 0.90 \\
Linear B & 0.0003 ($\pm$ 0.14) & 0.003 & 0.99 \\
Mandaic & -0.07 ($\pm$ 0.14) & -0.45 & 0.65 \\
Meetei Mayek & 0.01 ($\pm$ 0.13) & 0.04 & 0.97 \\
Meroitic Cursive & 0.10 ($\pm$ 0.19) & 0.54 & 0.59 \\
Mro & -0.16 ($\pm$ 0.15) & -1.06 & 0.29 \\
Multani & 0.01 ($\pm$ 0.14) & 0.06 & 0.95 \\
Nabataean & -0.02 ($\pm$ 0.15) & -0.12 & 0.90 \\
New Tai Lue & -0.03 ($\pm$ 0.15) & -0.21 & 0.84 \\
Nyiakeng Puachue Hmong& -0.21 ($\pm$ 0.15) & -1.40 & 0.16 \\
Ogham & -0.12 ($\pm$ 0.16) & -0.76 & 0.45 \\
Ol Chiki & -0.06 ($\pm$ 0.15) & -0.43 & 0.67 \\
Old Permic & -0.12 ($\pm$ 0.13) & -0.89 & 0.38 \\
Osmanya & -0.26 ($\pm$ 0.22) & -1.20 & 0.23 \\
Pau Cin Hau & -0.06 ($\pm$ 0.13) & -0.48 & 0.63 \\
Phoenician & -0.14 ($\pm$ 0.16) & -0.86 & 0.39 \\
Pracalit & 0.01 ($\pm$ 0.14) & 0.06 & 0.95 \\
Psalter Pahlavi & 0.01 ($\pm$ 0.16) & 0.09 & 0.93 \\
Runic & -0.03 ($\pm$ 0.13) & -0.24 & 0.81 \\
Shavian & -0.02 ($\pm$ 0.14) & -0.17 & 0.86 \\
Sora Sompeng & -0.14 ($\pm$ 0.14) & -1.06 & 0.29 \\
Soyombo & -0.12 ($\pm$ 0.13) & -0.90 & 0.37 \\
Syloti Nagri & -0.10 ($\pm$ 0.14) & -0.68 & 0.50 \\
Tagbanwa & -0.14 ($\pm$ 0.16) & -0.90 & 0.37 \\
Tai Viet & -0.10 ($\pm$ 0.15) & -0.68 & 0.50 \\
Warang Citi & -0.01 ($\pm$ 0.15) & -0.01 & 0.99 \\
Zanabazar Square & -0.08 ($\pm$ 0.13) & -0.64 & 0.52 \\
\hline
\end{tabular}
\caption{Full logistic regression table for \blip, language effects.}\label{tab:mem_blip2}
\end{table}

\clearpage

\begin{table}[t]
\begin{tabular}{|l|lll|}\hline
Fixed effects&$\widehat{\beta}$&$z$&$p$\\\hline
(Intercept) & 0.21 ($\pm$ 0.15) & 1.47 & 0.14 \\
color & 0.04 ($\pm$ 0.03) & 1.33 & 0.18 \\
leftright & 0.04 ($\pm$ 0.03) & 1.23 & 0.22 \\
length & 0.004 ($\pm$ 0.001) & 4.77 & \textbf{\textless 0.001} \\\hline
cross & -0.14 ($\pm$ 0.15) & -0.96 & 0.34 \\
diagonal & -0.02 ($\pm$ 0.15) & -0.14 & 0.89 \\
enclosed & 0.15 ($\pm$ 0.06) & 2.62 & \textbf{0.009} \\
horizontal & -0.19 ($\pm$ 0.10) & -1.89 & 0.060 \\
mirror & 0.45 ($\pm$ 0.27) & 1.70 & 0.089 \\
specific & 0.49 ($\pm$ 0.06) & 7.96 & \textbf{\textless 0.001} \\
stroke & -0.13 ($\pm$ 0.08) & -1.59 & 0.11 \\
symmetry & 0.01 ($\pm$ 0.08) & 0.18 & 0.85 \\
vertical & 0.14 ($\pm$ 0.10) & 1.37 & 0.17 \\\hline
\end{tabular}
\caption{Full logistic regression table for \molmoo, main effects.}\label{tab:mem_molmoo}
\end{table}

\begin{table}[t]
\begin{tabular}{|l|lll|}\hline
Fixed effects&$\widehat{\beta}$&$z$&$p$\\\hline
(Intercept) & 0.070 ($\pm$ 0.14) & 0.49 & 0.63 \\
color & 0.05 ($\pm$ 0.03) & 1.85 & 0.065 \\
leftright & 0.07 ($\pm$ 0.03) & 2.50 & 0.013 \\
length & 0.005 ($\pm$ \textless 0.001) & 5.84 & \textbf{\textless 0.001} \\\hline
cross & -0.20 ($\pm$ 0.14) & -1.36 & 0.18 \\
diagonal & 0.35 ($\pm$ 0.15) & 2.37 & \textbf{0.02} \\
enclosed & 0.003 ($\pm$ 0.06) & 0.06 & 0.95 \\
horizontal & -0.19 ($\pm$ 0.10) & -1.90 & 0.059 \\
mirror & 0.62 ($\pm$ 0.27) & 2.35 & \textbf{0.02} \\
specific & 0.45 ($\pm$ 0.06) & 7.29 & \textbf{\textless 0.001} \\
stroke & -0.18 ($\pm$ 0.08) & -2.24 & \textbf{0.03} \\
symmetry & -0.29 ($\pm$ 0.08) & -3.64 & \textbf{\textless 0.001} \\
vertical & 0.23 ($\pm$ 0.10) & 2.18 & \textbf{0.03} \\\hline
\end{tabular}
\caption{Full logistic regression table for \molmod, main effects.}\label{tab:mem_molmod}
\end{table}

\clearpage

\begin{table}[t]
\begin{tabular}{|l|lll|}\hline
Fixed effects&$\widehat{\beta}$&$z$&$p$\\\hline
Avestan & -0.25 ($\pm$ 0.18) & -1.36 & 0.18 \\
Bamum & -0.13 ($\pm$ 0.20) & -0.67 & 0.50 \\
Bassa Vah & -0.05 ($\pm$ 0.18) & -0.26 & 0.79 \\
Batak & 0.01 ($\pm$ 0.18) & 0.05 & 0.96 \\
Buginese & 0.14 ($\pm$ 0.23) & 0.61 & 0.54 \\
Cherokee & -0.05 ($\pm$ 0.17) & -0.31 & 0.76 \\
Gothic & -0.18 ($\pm$ 0.21) & -0.84 & 0.40 \\
Grantha & -0.32 ($\pm$ 0.16) & -1.97 & 0.05 \\
Hangul & 0.06 ($\pm$ 0.18) & 0.32 & 0.75 \\
Kayah Li & -0.21 ($\pm$ 0.17) & -1.19 & 0.24 \\
Khudawadi & -0.32 ($\pm$ 0.21) & -1.51 & 0.13 \\
Lao & -0.10 ($\pm$ 0.17) & -0.61 & 0.54 \\
Lepcha & 0.08 ($\pm$ 0.18) & 0.44 & 0.66 \\
Linear B & -0.10 ($\pm$ 0.17) & -0.57 & 0.57 \\
Mandaic & -0.14 ($\pm$ 0.18) & -0.77 & 0.44 \\
Meetei Mayek & -0.09 ($\pm$ 0.16) & -0.57 & 0.57 \\
Meroitic Cursive & -0.09 ($\pm$ 0.23) & -0.37 & 0.71 \\
Mro & -0.32 ($\pm$ 0.19) & -1.71 & 0.09 \\
Multani & 0.01 ($\pm$ 0.17) & 0.08 & 0.94 \\
Nabataean & -0.05 ($\pm$ 0.18) & -0.30 & 0.76 \\
New Tai Lue & 0.20 ($\pm$ 0.19) & 1.04 & 0.30 \\
Nyiakeng Puachue Hmong & -0.05 ($\pm$ 0.18) & -0.28 & 0.78 \\
Ogham & 0.05 ($\pm$ 0.20) & 0.25 & 0.80 \\
Ol Chiki & 0.11 ($\pm$ 0.18) & 0.58 & 0.56 \\
Old Permic & -0.28 ($\pm$ 0.17) & -1.66 & 0.10 \\
Osmanya & 0.15 ($\pm$ 0.27) & 0.58 & 0.56 \\
Pau Cin Hau & -0.25 ($\pm$ 0.16) & -1.55 & 0.12 \\
Phoenician & -0.15 ($\pm$ 0.20) & -0.77 & 0.44 \\
Pracalit & -0.18 ($\pm$ 0.17) & -1.05 & 0.29 \\
Psalter Pahlavi & -0.04 ($\pm$ 0.20) & -0.22 & 0.83 \\
Runic & -0.12 ($\pm$ 0.16) & -0.71 & 0.48 \\
Shavian & -0.12 ($\pm$ 0.18) & -0.70 & 0.48 \\
Sora Sompeng & -0.16 ($\pm$ 0.17) & -0.95 & 0.34 \\
Soyombo & -0.02 ($\pm$ 0.16) & -0.10 & 0.92 \\
Syloti Nagri & 0.04 ($\pm$ 0.17) & 0.22 & 0.82 \\
Tagbanwa & -0.02 ($\pm$ 0.20) & -0.10 & 0.92 \\
Tai Viet & -0.05 ($\pm$ 0.18) & -0.28 & 0.78 \\
Warang Citi & -0.10 ($\pm$ 0.19) & -0.52 & 0.61 \\
Zanabazar Square & -0.003 ($\pm$ 0.16) & -0.02 & 0.99 \\
\hline
\end{tabular}
\caption{Full logistic regression table for \molmoo, language efffects.}\label{tab:mem_molmoo2}
\end{table}

\clearpage

\begin{table}[t]
\begin{tabular}{|l|lll|}\hline
Fixed effects&$\widehat{\beta}$&$z$&$p$\\\hline
Avestan & 0.02 ($\pm$ 0.18) & 0.08 & 0.93 \\
Bamum & 0.05 ($\pm$ 0.20) & 0.24 & 0.81 \\
Bassa Vah & 0.06 ($\pm$ 0.18) & 0.34 & 0.73 \\
Batak & 0.28 ($\pm$ 0.18) & 1.56 & 0.12 \\
Buginese & 0.68 ($\pm$ 0.23) & 2.94 & \textbf{0.004} \\
Cherokee & 0.14 ($\pm$ 0.17) & 0.80 & 0.42 \\
Gothic & 0.18 ($\pm$ 0.21) & 0.84 & 0.40 \\
Grantha & -0.10 ($\pm$ 0.16) & -0.59 & 0.55 \\
Hangul & 0.22 ($\pm$ 0.18) & 1.25 & 0.21 \\
Kayah Li & 0.08 ($\pm$ 0.17) & 0.47 & 0.64 \\
Khudawadi & 0.07 ($\pm$ 0.21) & 0.33 & 0.74 \\
Lao & 0.05 ($\pm$ 0.17) & 0.28 & 0.78 \\
Lepcha & 0.04 ($\pm$ 0.18) & 0.21 & 0.83 \\
Linear B & 0.17 ($\pm$ 0.17) & 0.98 & 0.33 \\
Mandaic & 0.03 ($\pm$ 0.18) & 0.17 & 0.87 \\
Meetei Mayek & 0.003 ($\pm$ 0.16) & 0.02 & 0.99 \\
Meroitic Cursive & 0.13 ($\pm$ 0.23) & 0.55 & 0.58 \\
Mro & -0.10 ($\pm$ 0.19) & -0.51 & 0.61 \\
Multani & 0.08 ($\pm$ 0.17) & 0.50 & 0.62 \\
Nabataean & -0.04 ($\pm$ 0.18) & -0.22 & 0.83 \\
New Tai Lue & 0.18 ($\pm$ 0.19) & 0.95 & 0.34 \\
Nyiakeng Puachue Hmong & 0.10 ($\pm$ 0.18) & 0.56 & 0.57 \\
Ogham & 0.08 ($\pm$ 0.20) & 0.43 & 0.67 \\
Ol Chiki & 0.17 ($\pm$ 0.18) & 0.91 & 0.36 \\
Old Permic & 0.16 ($\pm$ 0.17) & 0.99 & 0.32 \\
Osmanya & -0.08 ($\pm$ 0.27) & -0.29 & 0.78 \\
Pau Cin Hau & 0.01 ($\pm$ 0.16) & 0.08 & 0.94 \\
Phoenician & -0.03 ($\pm$ 0.20) & -0.13 & 0.90 \\
Pracalit & 0.005 ($\pm$ 0.17) & 0.03 & 0.98 \\
Psalter Pahlavi & 0.29 ($\pm$ 0.20) & 1.47 & 0.14 \\
Runic & 0.03 ($\pm$ 0.16) & 0.20 & 0.84 \\
Shavian & -0.03 ($\pm$ 0.18) & -0.15 & 0.88 \\
Sora Sompeng & -0.07 ($\pm$ 0.17) & -0.39 & 0.69 \\
Soyombo & -0.06 ($\pm$ 0.16) & -0.35 & 0.73 \\
Syloti Nagri & -0.006 ($\pm$ 0.17) & -0.03 & 0.97 \\
Tagbanwa & 0.44 ($\pm$ 0.20) & 2.25 & \textbf{0.026} \\
Tai Viet & 0.02 ($\pm$ 0.18) & 0.08 & 0.93 \\
Warang Citi & 0.24 ($\pm$ 0.19) & 1.28 & 0.20 \\
Zanabazar Square & 0.21 ($\pm$ 0.16) & 1.31 & 0.19 \\
\hline
\end{tabular}
\caption{Full logistic regression table for \molmod, language effects.}\label{tab:mem_molmod2}
\end{table}

\end{document}